\documentclass[letterpaper]{article} 
\usepackage{aaai23}  
\usepackage{times}  
\usepackage{helvet}  
\usepackage{courier}  
\usepackage[hyphens]{url}  
\usepackage{graphicx} 
\usepackage{natbib}  
\usepackage{caption} 
\usepackage{algorithm}
\usepackage{algorithmic}
\usepackage{subcaption}
\usepackage{amsmath}
\usepackage{amsthm}
\usepackage{amssymb}
\usepackage{paralist}
\usepackage{dsfont}

\theoremstyle{plain}
\newtheorem{theorem}{Theorem}[section]

\theoremstyle{definition}
\newtheorem{definition}[theorem]{Definition}

\theoremstyle{remark}

\usepackage{newfloat}
\usepackage{listings}
\DeclareCaptionStyle{ruled}{labelfont=normalfont,labelsep=colon,strut=off} 
\floatstyle{ruled}
\newfloat{listing}{tb}{lst}{}
\floatname{listing}{Listing}
\title{Towards Robust Metrics For Concept Representation Evaluation}
\author {
    Mateo Espinosa Zarlenga\textsuperscript{\rm 1}\equalcontrib,
    Pietro Barbiero\textsuperscript{\rm 1}\equalcontrib, 
    Zohreh Shams\textsuperscript{\rm 1, \rm 2}\equalcontrib,
    Dmitry Kazhdan\textsuperscript{\rm 1}, 
    Umang Bhatt\textsuperscript{\rm 1, \rm 3}, 
    Adrian Weller\textsuperscript{\rm 1, \rm 3}, 
    Mateja Jamnik\textsuperscript{\rm 1}
}
\affiliations {
    \textsuperscript{\rm 1} University of Cambridge\\
    \textsuperscript{\rm 2} Babylon Health\\
    \textsuperscript{\rm 3} The Alan Turing Institute\\
    \{me466, pb737, zs315, dk525, usb20, aw665\}@cam.ac.uk, mateja.jamnik@cl.cam.ac.uk
}
\usepackage{bibentry}

\begin{document}

\maketitle

\begin{abstract}
Recent work on interpretability has focused on concept-based explanations, where deep learning models are explained in terms of high-level units of information, referred to as concepts. Concept learning models, however, have been shown to be prone to encoding impurities in their representations, failing to fully capture meaningful features of their inputs. While concept learning lacks metrics to measure such phenomena, the field of disentanglement learning has explored the related notion of underlying factors of variation in the data, with plenty of metrics to measure the purity of such factors. In this paper, we show that such metrics are not appropriate for concept learning and propose novel metrics for evaluating the purity of concept representations in both approaches. We show the advantage of these metrics over existing ones and demonstrate their utility in evaluating the robustness of concept representations and interventions performed on them. In addition, we show their utility for benchmarking state-of-the-art methods from both families and find that, contrary to common assumptions, supervision alone may not be sufficient for pure concept representations.
\end{abstract}

\section{Introduction}
Addressing the lack of interpretability of deep neural networks (DNNs) has given rise to explainability methods, most common of which are feature importance methods~\citep{Riberio2016,Lundberg2017} that quantify the contribution of input features to certain predictions~\citep{bhatt2020explainable}. However, input features may not necessarily form the most intuitive basis for explanations, in particular when using low-level features such as pixels. \emph{Concept-based explainability}~\citep{Been2018, Ghorbani2019, Koh2020, Been2020, ciravegna2021logic} remedies this issue by constructing an explanation at a concept level, where concepts are considered intermediate, high-level and semantically meaningful units of information commonly used by humans to explain their decisions. Recent work, however, has shown that concept learning (CL) models may not correctly capture the intended semantics of their representations~\cite{Margeloiu2021}, and that their learnt concept representations are prone to encoding  \emph{impurities} (i.e., more information in a concept than what is intended)~\citep{Mahin2021}. Such phenomena may have severe consequences for how such representations can be interpreted (as shown in the misleading attribution maps described by~\citet{Margeloiu2021}) and used in practice (as shown later in our intervention results). Nevertheless, the CL literature is yet to see concrete metrics to appropriately capture and measure these phenomena.

In contrast, the closely-related field of \emph{disentanglement learning} (DGL)~\citep{Bengio2013, Higgins2017, Locatello2019, Locatello2020b}, where methods aim to learn intermediate representations aligned to disentangled factors of variation in the data, offers a wide array of metrics for evaluating the quality of latent representations. However, despite the close relationship between concept representations in CL and latent codes in DGL, metrics proposed in DGL are built on assumptions that do not hold in CL, as explained in Section~\ref{sec:background}, and are thus inappropriate to measure the aforementioned undesired phenomena in CL.

In this paper, we show the inadequacy of current metrics and introduce two novel metrics for evaluating the purity of intermediate representations in CL. Our results indicate that our metrics can be used in practice for quality assurance of such intermediate representations for:
\begin{compactenum} 
\item Detecting impurities (i) concealed in soft representations, (ii) caused by different model capacities, or (iii) caused by spurious correlations.
\item Indicating when concept interventions are safe.
\item Revealing the impact of supervisions on concept purity.
\item Being robust to inter-concept correlations.
\end{compactenum} 

\section{Background and Motivation}
\label{sec:background}
\paragraph{Notation} 
In CL, the aim is to find a low-dimensional intermediate representation $\mathbf{\hat{c}}$ of the data, 
similar to latent codes $\mathbf{\hat{z}}$ in DGL. This low-dimensional representation corresponds to a matrix $\mathbf{\hat{c}} \in \hat{C} \subseteq \mathbb{R}^{d \times k}$ in which the $i$-th column constitutes a $d$-dimensional representation of the $i$-th concept, assuming that the length of all concept representations can be made equal using zero-padding. Under this view, elements in $\mathbf{\hat{c}}_{(:, i)} \in \mathbb{R}^d$ are expected to have high values (under some reasonable aggregation function) if the $i$-th concept is considered to be activated. As most CL methods assume $d=1$, for succinctness we use $\hat{c}_i$ in place of $\mathbf{\hat{c}}_{(:, i)}$ when $d = 1$. Analogously, as each latent code  $\mathbf{\hat{z}}_{(:, i)}$ aims to encode an independent factor of variation (i.e., concept) $z_i$ in DGL, we use $\mathbf{\hat{c}}$ for both learnt concept representations and latent codes. Similarly, we refer to both ground truth concepts and factors of variations as $\mathbf{c} \in C \subseteq \mathbb{R}^{k}$.

\paragraph{Concept Learning}
In supervised CL, access to concept labels for each input, in addition to task labels, is assumed. Supervised CL makes use of: (i) a concept encoder function $g: X \mapsto \hat{C}$ that maps the inputs to a concept representation; and (ii) a label predictor function $f: \hat{C} \mapsto Y$ that maps the concept representations to a downstream task's set of labels $\mathbf{y} \in Y \subseteq \mathbb{R}^L$. Together, these two functions constitute a \textit{Concept Bottleneck Model} (CBM)~\citep{Koh2020}. A notable approach that uses the bottleneck idea is \textit{Concept Whitening} (CW)~\citep{Chen2020} which introduces a batch normalisation module whose activations are trained to be aligned with sets of binary concepts. Unlike supervised CL, in unsupervised CL concept annotations are not available and concepts are discovered in an unsupervised manner with the help of task labels. Two notable data modality agnostic approaches in this family are \textit{Completeness-aware Concept Discovery} (CCD)~\citep{Been2020} and \textit{Self-Explainable Neural Networks} (SENNs)~\citep{alvarez2018towards}. We refer to supervision from ground truth concepts in supervised CL as \emph{explicit}, while supervision from task labels alone is referred to as \emph{implicit}.

On the other hand, generative models (e.g., VAEs~\citep{Kingma2014}) used in DGL assume that data is generated from a set of independent factors of variation $\mathbf{c} \in C$. Thus, the goal is to find a function $g(\cdot)$ that maps inputs to a disentangled latent representation. In the light of recent work~\citep{Locatello2019} showing the impossibility of learning disentangled representations without any supervision, as in $\beta$-VAEs~\citep{Higgins2017}, recent work suggests using \emph{weak} supervision for learning latent codes~\citep{Locatello2020}.

\begin{figure}[t!]
    \centering
    \begin{subfigure}[b]{0.4\columnwidth}
    \centering
        \includegraphics[width=0.9\textwidth]{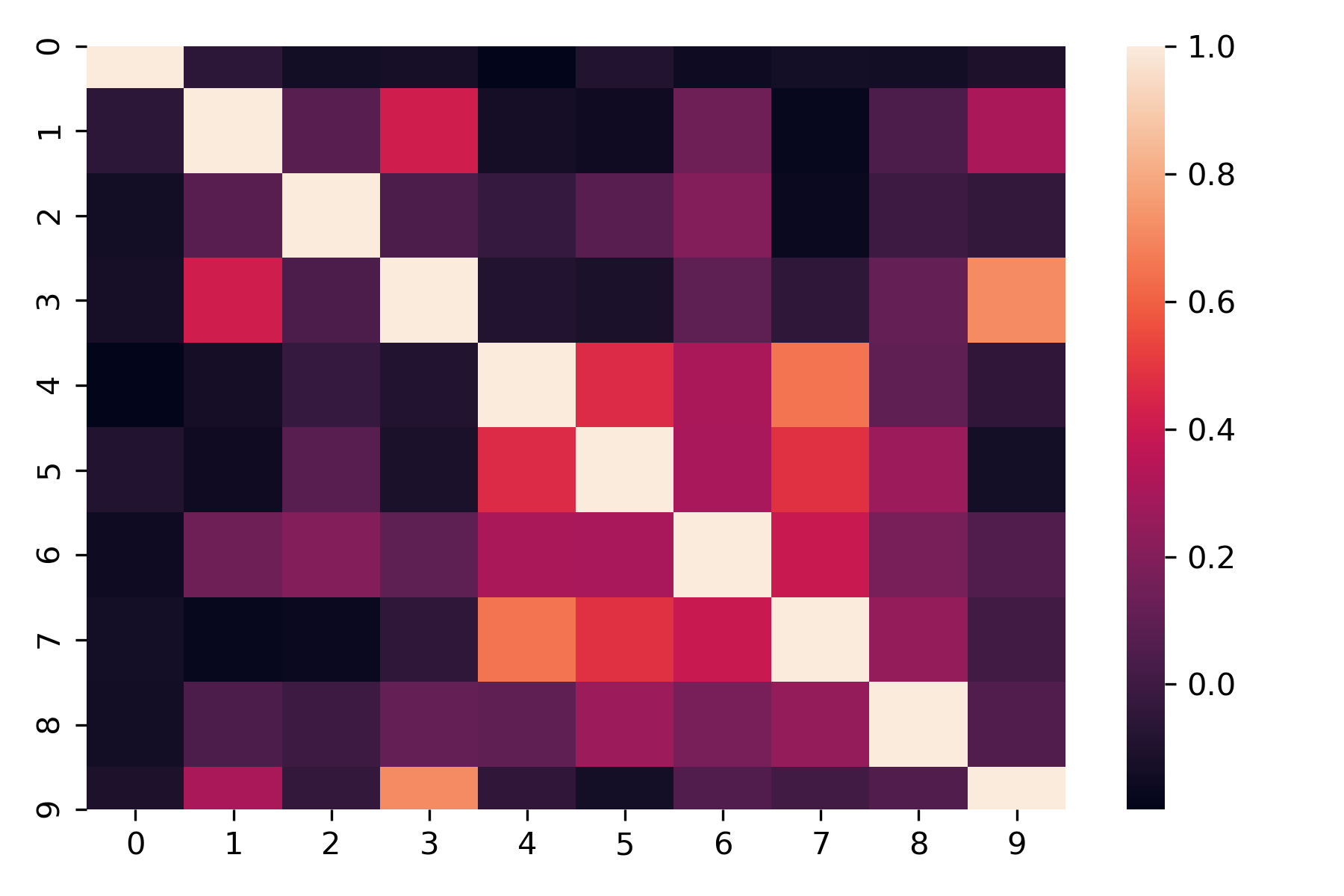}
        \caption{}
        \label{fig:CUB}
    \end{subfigure}
    \begin{subfigure}[b]{0.55\columnwidth}
        \centering
        \includegraphics[width=0.9\textwidth]{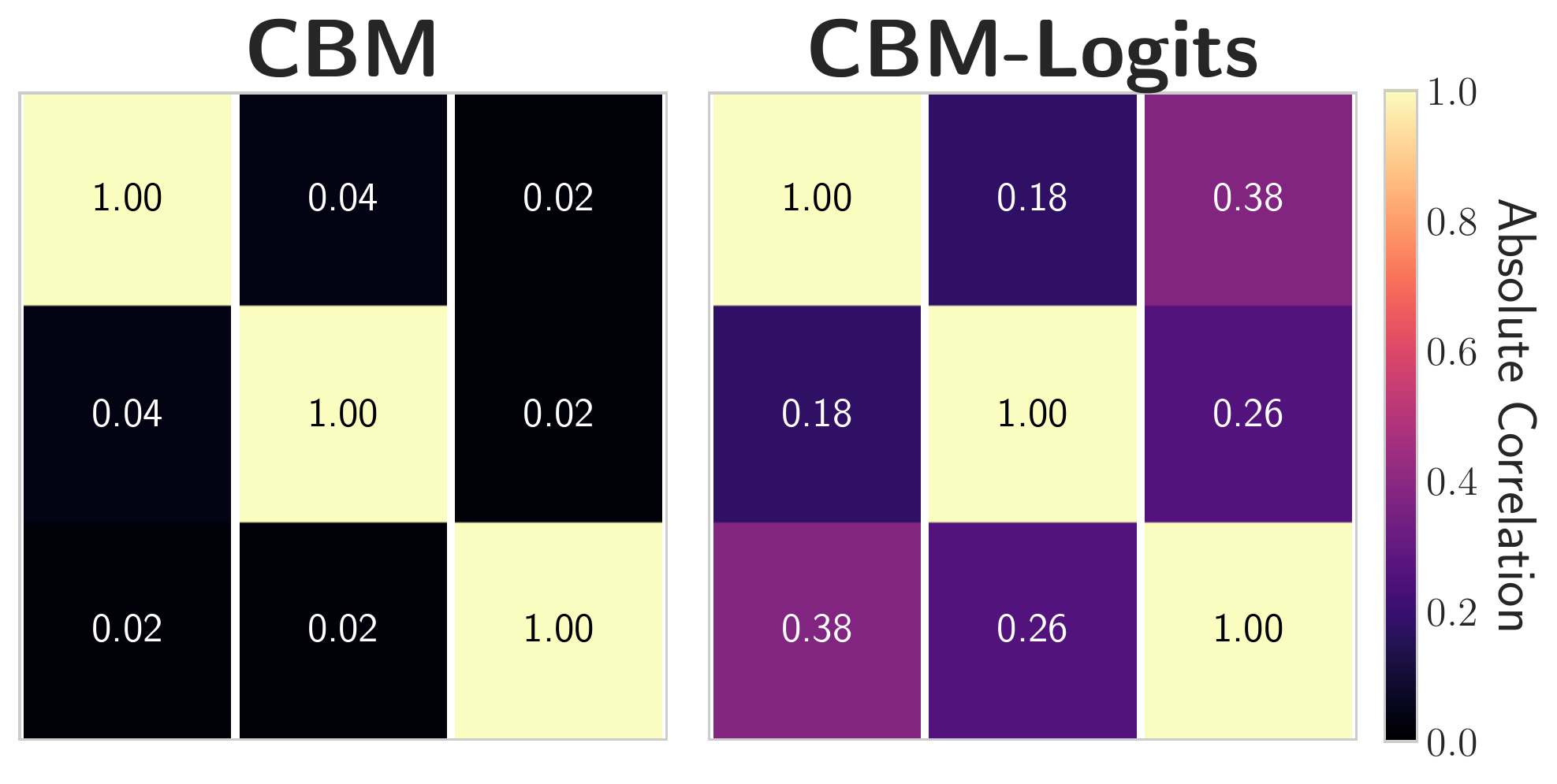}
        \caption{}
        \label{fig:SameAccuracy}
    \end{subfigure}
    \caption{(a) Absolute correlation of the top-10 concepts with highest label mutual information in CUB dataset.  (b) Two identically-trained CBMs with almost identical accuracies yet different levels of inter-concept correlations.}
\end{figure}

\paragraph{Shortcomings of Current Metrics}
Generally, the DGL literature defines concept quality in terms of disentanglement i.e., the more learnt concepts are decorrelated the better (see Appendix~\ref{appendix:metricReview} for a summary of DGL metrics). We argue that existing DGL metrics are inadequate to ensure concept quality in CL as they: (i) Assume that each concept is represented with a single scalar value, which is not the case in some modern CL methods such as CW. (ii) Fail to capture subtle impurities encoded within continuous representations (as demonstrated in Section~\ref{sec:experiments}). (iii) May assume access to a tractable concept-to-sample generative process (something uncommon in real-world datasets). (iv) Assume that inter-concept dependencies are undesired, an assumption that may not be realistic in the real world where ground truth concept labels often are correlated. This can be  observed in Figure~\ref{fig:CUB} where concept labels in the Caltech-UCSD Birds-200-2011 (CUB) dataset~\citep{CUB}, a widely used CL benchmark, are seen to be \textit{highly correlated}.

Metrics in CL~\citep{Koh2020, Been2020, Kazhdan2020}, on the other hand, mainly define concept quality w.r.t.\ the downstream task (e.g., \emph{task predictive accuracy}), and rarely evaluate properties of concept representations w.r.t.\ the ground truth concepts (with the exception of \emph{concept predictive accuracy}). Nevertheless, it is possible for two CL models to learn concept representations that yield similar task and concept accuracy but have vastly different properties/qualities. For example, in Figure~\ref{fig:SameAccuracy} we show a simple experiment in which two CBMs trained on a toy dataset with 3 independent concepts, where ``\textit{CBM}'' uses a sigmoidal bottleneck and ``\textit{CBM-Logits}'' uses plain logits in its bottleneck, generate concept representations with the same concept/task accuracies yet significantly different inter-concept correlations (details in Appendix~\ref{appendix:correlation_toy_experiment}).

\section{Measuring Purity of Concept Representations}
To address the shortcomings of existing metrics, in this section we propose two metrics that make no assumptions about (i) the presence/absence of correlations between concepts, (ii) the underlying data-generating process, and (iii) the dimensionality of a concept representation. Specifically, we focus on measuring the quality of a concept representation in terms of their ``\textit{purity}'', defined here as whether the predictive power of a learnt concept representation over other concepts is similar to what we would expect from their corresponding ground truth labels. We begin by introducing the \textbf{oracle impurity score} (OIS), a metric that quantifies impurities \textit{localised} within individual learnt concepts. Then, we introduce the \textbf{niche impurity score} (NIP) as a metric that focuses on capturing impurities \textit{distributed} across the set of learnt concept representations. 

\subsection{Oracle Impurity}
To circumvent the aforementioned limitations of existing DGL metrics, we take inspiration from~\citep{Mahin2021}, where they informally measure concept impurity as how predictive a CBM-generated concept probability is for the ground truth value of other independent concepts. If the pre-defined concepts are independent, then the inter-concept predictive performance should be no better than random. To generalise this assumption beyond independent concepts, we first measure the predictability of ground truth concepts w.r.t.\ one another. Then we measure the predictability of learnt concepts w.r.t.\ the ground truth ones. The divergence between the former and the latter acts as an impurity metric, measuring the amount of undesired information that is encoded, or lacking, in the learnt concepts. To formally introduce our metric, we begin by defining a \textit{purity matrix}.

\begin{definition}[Purity Matrix]
\label{def:purity_matrix}
    Given a set of $n$ concept representations $\hat{\Gamma} = \{\mathbf{\hat{c}}^{(l)} \in \mathbb{R}^{d \times k}\}_{l = 1}^n$, and corresponding discrete ground truth concept annotations $\Gamma = \{\mathbf{c}^{(l)} \in \mathbb{N}^{k} \}_{l = 1}^n$, 
    assume that $\hat{\Gamma}$ and $\Gamma$ are aligned element-wise: for all $i \in \{1, \cdots, k\}$, the $i$-th concept representation of $\mathbf{\hat{c}}^{(l)}$ encodes the same concept as the $i$-th concept label in $\mathbf{c}^{(l)}$. The Purity Matrix of $\hat{\Gamma}$, given ground truth labels $\Gamma$, is defined as a matrix $\mathbf{\pi}\big(\hat{\Gamma}, \Gamma\big) \in [0, 1]^{k \times k}$ whose entries are given by:
    \[
        \mathbf{\pi}\big(\hat{\Gamma}, \Gamma\big)_{(i, j)} := \text{AUC-ROC}\big(\big\{ \big(\psi_{i, j} ( \mathbf{\hat{c}}^{(l)}_{(:,i)} \big), c^{(l)}_j \big)\big\}_{l=1}^n\big) ,
    \]
    where $\psi_{i, j}(\cdot)$ is a non-linear model (e.g., an MLP) mapping the $i$-th concept's representation $\mathbf{\hat{c}}_{(:,i)}$ to a probability distribution over all values concept $j$ may take.
\end{definition}

The $(i, j)$-th entry of $\mathbf{\pi}\big(\hat{\Gamma}, \Gamma\big)$ contains the AUC-ROC when predicting the ground truth value of concept $j$ given the $i$-th concept representation.
Therefore, the diagonal entries of this matrix show how good a concept representation is at predicting its aligned ground truth label, while the off-diagonal entries show how good such a representation is at predicting the ground truth labels of other concepts.
Intuitively, one can think of the $(i, j)$-th entry of this matrix as a proxy of the mutual information between the $i$-th concept representation and the $j$-th ground truth concept. While in principle other methods could be used to estimate this mutual information (e.g., histogram binning), we choose the test AUC-ROC of a trained non-linear model primarily for its tractability, its bounded nature, and its easy generalisation to non-scalar concept representations.
Furthermore, while in this work we focus on concepts that are binary in nature, our definition can be applied to multivariate concepts by using the mean one-vs-all AUC-ROC score. See Appendix~\ref{appendix:oracle_impurity_details} for implementation details and Appendix~\ref{appendix:metric_model_ablation} for a discussion on how the OIS is robust to the model selected for $\psi_{i, j}(\cdot)$.

This matrix allows us to construct a metric for quantifying the impurity of a concept encoder:

\begin{definition}[Oracle Impurity Score (OIS)]
\label{def:oracle_impurity}
    Let $g: X \mapsto \hat{C} \subseteq \mathbb{R}^{d \times k}$ be a concept encoder and let $ \Gamma_X := \{ \mathbf{x}^{(i)} \in X \}_{i = 1}^n$ and $\Gamma~:=~\{ \mathbf{c}^{(i)} \in \mathbb{N}^k\}_{i = 1}^n$ be ordered sets of testing samples and their corresponding concept annotations, respectively. If, for any ordered set $A$ we define $g(A)$ as $g(A) := \{g(a) \; | \; a \in A \}$, then the $\text{OIS}$
    is defined as:
    \begin{equation*}
        \text{OIS}(g, \Gamma_X, \Gamma) := \frac{2 \Big|\Big| \mathbf{\pi}\big(g(\Gamma_X), \Gamma\big) - \mathbf{\pi}\big(\Gamma, \Gamma\big) \Big|\Big|_F }{k}
    \end{equation*}
    where $||\mathbf{A}||_F$ represents the Frobenius norm of $\mathbf{A}$.
\end{definition}

Intuitively, the OIS measures the total deviation of an encoder's purity matrix with the purity matrix obtained from using the ground truth concept labels only (i.e., the ``oracle matrix''). We opt to measure this divergence using the Frobenius norm of their difference in order to obtain a bounded output which can be easily interpreted. Since each entry in the difference $\left(\mathbf{\pi}\big(g(\Gamma_X), \Gamma\big) - \mathbf{\pi}\big(\Gamma, \Gamma\big)\right)$ can be at most $1/2$, the upper bound of $\Big|\Big| \mathbf{\pi}\big(g(\Gamma_X), \Gamma\big) - \mathbf{\pi}\big(\Gamma, \Gamma\big) \Big|\Big|_F$
is $k/2$. Therefore, the OIS includes a factor of $2/k$ to guarantee that it ranges in $[0, 1]$. This allows interpreting an OIS of $1$ as a complete misalignment between $\mathbf{\pi}\big(\Gamma, \Gamma\big)$ and  $\mathbf{\pi}\big(g(\Gamma_X), \Gamma\big)$ (i.e., the $i$-th concept representation can predict all other concept labels except its corresponding one even when concepts are independent). An impurity score of $0$, on the other hand, represents a perfect alignment between the two purity matrices (i.e., the $i$-th concept representation does not encode any unnecessary information for predicting  concept $i$).

\subsection{Niche Impurity} \label{sec:niching}
While the OIS is able to correctly capture impurities that are localised within specific and individual concept representations, it is also possible that information pertinent to unrelated concepts is encoded \textit{across multiple learnt representations}. To tractably capture such a phenomenon, we propose the \textit{Niching Impurity Score} (NIS) inspired by the theory of niching. In ecology, a niche is considered to be a resource-constrained subspace of the environment that can support different types of life \citep{darwin1859origin}. Analogously, the NIS looks at the predictive power of subsets of disentangled concepts. In contrast with the OIS, the NIS is concerned with impurities encoded in \textit{sets} of learnt concept representations rather than impurities in individual concept representations. The NIS efficiently quantifies the amount of shared information across concept representations by looking at how predictive disentangled subsets of concept representations are for ground truth concepts. We start by describing a \emph{concept nicher}, a function that ranks learnt concepts by how much information they share with the ground truth ones. We then define a \emph{concept niche} for a ground truth concept as a set of learnt concepts that are highly ranked by the concept nicher, while the set of concepts outside the niche is referred to as the \emph{concept niche complement}. We conclude by constructing the NIS by looking at how predictable a ground truth concept is from its corresponding concept niche complement. The collective NIS of all concepts, therefore, represents impurities encoded across the entire bottleneck.

\begin{definition}[Concept nicher] \label{def:nicher}
Given a set of concept representations $\hat{C} \subseteq \mathbb{R}^{d \times k}$, we define a concept nicher as a function $\nu: \{1, \cdots k\} \times \{1, \cdots k\} \mapsto [0, 1]$ that returns $\nu(i, j) \approx 1$ if the $i$-th concept $\mathbf{\hat{c}}_{(:, i)}$ is entangled with the $j$-th ground truth concept $c_j$, and $\nu(i, j) \approx 0$ otherwise.
\end{definition}

Our definition above can be instantiated in various ways, depending on how entanglement is measured. In favour of efficiency, we measure entanglement using absolute Pearson correlation $\rho$, as this measure can efficiently discover (a linear form of) association between variables~\cite{altman2015points}. We call this instantiation  \emph{concept-correlation nicher} (CCorrN) and define it as
$\text{CCorrN}(i, j) := \big| \rho\big(\{\mathbf{\hat{c}}^{(l)}_{(:, i)}\}_{l=1}^N, \{c^{(l)}_j\}_{l=1}^N\big) \big|$.

If $\mathbf{\hat{c}}_{(:, i)}$ is not a scalar representation (i.e., $d > 1$), then for simplicity, we use the maximum absolute correlation coefficient between all entries in $\mathbf{\hat{c}}_{(:, i)}$, and the target concept label $c_j$ as a representative correlation coefficient for the entire representation $\mathbf{\hat{c}}_{(:, i)}$. We then define a concept niche as: 
\begin{definition}[Concept niche]
The concept niche $N_j(\nu, \beta)$ for target concept $j$, determined by concept nicher $\nu(\cdot, \cdot)$ and threshold $\beta \in [0,1]$, is defined as $N_j(\nu, \beta) := \big\{i \ \ | \ \ i \in \{1, \cdots, k\} \text{ and } \nu(i, j) > \beta \big\}$.
\end{definition}

From this, the Niche Impurity (NI) measures the predictive capacity of the complement of concept niche $N_i(\nu, \beta)$, referred to as $\neg N_i(\nu, \beta) := \{1, \cdots, k\} \; \backslash \; N_i(\nu, \beta)$, for the $i$-th ground truth concept:

\begin{definition}[Niche Impurity (NI)] \label{def:niche_impurity}
Given a classifier $f: \hat{C} \mapsto C$, concept nicher $\nu$, threshold $\beta \in [0, 1]$, and labeled concept representations $\{(\mathbf{\hat{c}}^{(l)}, \mathbf{c}^{(l)})\}_{l = 1}^n$, the Niche Impurity of the $i$-th output of $f(\cdot)$ is defined as $\text{NI}_i(f, \nu, \beta) := \text{AUC-ROC} \big( \{( f|_{\neg N_i(\nu, \beta)} \big( \mathbf{\hat{c}}^{(l)}_{(:, \neg N_i(\nu, \beta))} \big), c^{(l)}_i) \}_{l=1}^n \big)$, where $f|_{\neg N_j(\nu, \beta)}$ is the classifier resulting from masking all entries in $\neg N_j(\nu, \beta)$ when feeding $f$ with concept representations. 
\end{definition}

Although $f$ can be any classifier, for simplicity in our experiments we use a ReLU MLP with hidden layer sizes $\{ 20, 20 \}$ (see Appendix~\ref{appendix:metric_model_ablation} for a discussion on our metric's robustness to $f$'s architecture).
Intuitively, a NI of $1/2$ (random AUC of niche complement) indicates that the concepts inside the niche $N_i(\nu)$ are the only concepts predictive of the $i$-th concept, that is, concepts outside the niche do not hold any predictive information of the $i$-th concept. Finally, the \textit{Niche Impurity Score} metric measures how much information apparently disentangled concepts are actually sharing:

\begin{definition}[Niche Impurity Score (NIS)] \label{def:niche_impurity_score}
Given a classifier $f: \hat{C} \mapsto C$ and concept nicher $\nu$, the niche impurity score $\text{NIS}(f,\nu) \in [0,1]$ is defined as the summation of niche impurities across all concepts for different values of $\beta$: $\text{NIS}(f,\nu) := \int_{0}^{1} (\sum_{i=1}^{k} \text{NI}_i(f, \nu, \beta)/k) d\beta$.
\end{definition}

In practice, this integral is estimated using the trapezoid method with $\beta \in \{ 0.0, 0.05, \cdots, 1\}$. Furthermore, because we parameterise $f$ as a small MLP, leading to a tractable impurity metric that scales with large concept sets. Intuitively, a NIS of $1$ means that all the information to perfectly predict each ground truth concept is spread on many different and disentangled concept representations. In contrast, a NIS around $1/2$ (random AUC) indicates that no concept can be predicted by any concept representation subset.

\section{Experiments}
\label{sec:experiments}
We now give a brief account of the experimental setup and datasets, followed by highlighting the utility of our impurity metrics and their applications to model benchmarking. 

\paragraph{Datasets} To have datasets compatible with both CL and DGL, we construct tasks whose samples are fully described by a vector of ground truth generative factors. Moreover, we simulate real-world scenarios by designing tasks with varying degrees of dependencies in their concept annotations. To achieve this, we first design a parametric binary-class dataset $\emph{TabularToy}(\delta)$, a variation of the tabular dataset proposed by~\citet{Mahin2021}. We also construct two multiclass image-based parametric datasets: $\emph{dSprites}(\lambda)$ and $\emph{3dshapes}(\lambda)$, based on dSprites \citep{matthey2017dSprites} and 3dshapes \citep{3dshapes18} datasets, respectively. They consist of 3D samples generated from a vector consisting of $k=5$ and $k=6$ factors of variation, respectively. Both datasets have one binary concept annotation per factor of variation. Parameters $\delta \in [0, 1]$ and $\lambda \in \{0, \cdots, k - 1\}$ control the degree of concept inter-dependencies during generation: $\lambda=0$ and $\delta = 0$ represent inter-concept independence while higher values represent stronger inter-concept dependencies. For dataset details see Appendix~\ref{appendix:dataset_details}. 

\begin{table*}[t]
    \centering
    \begin{tabular}{c|cc|cccc}
        {} &
            OIS ($\downarrow$) &
            NIS ($\downarrow$) &
            SAP ($\uparrow$) &
            MIG ($\uparrow$) &
            $R^4$ ($\uparrow$) &
            DCI Dis ($\uparrow$) \\ \hline
        Baseline Soft (\%) &
            \textbf{4.69 $\pm$ 0.43} &
            \textbf{66.25 $\pm$ 2.31} &
            48.74 $\pm$ 0.41 &
            99.93 $\pm$ 0.03 &
            99.95 $\pm$ 0.00 &
            \textbf{99.99 $\pm$ 0.00} \\
        Impure Soft (\%) &
            22.58 $\pm$ 2.34 &
            72.36 $\pm$ 1.26 &
            48.83 $\pm$ 0.53 &
            99.93 $\pm$ 0.04 &
            99.95 $\pm$ 0.00 &
            99.50 $\pm$ 0.01 \\
        $p$-value &
            $7.38 \times 10^{-5}$ &
            $3.24 \times 10^{-3}$ &
            $7.89 \times 10^{-1}$ &
            $9.26 \times 10^{-1}$ &
            $9.76 \times 10^{-1}$ &
            $3.66 \times 10^{-9}$ \\
            \hline
    \end{tabular}
    \caption{Comparison between our metrics (left of the middle line) and common DGL metrics (right) using hand-crafted soft concept representations and latent codes. We highlight statistically significant differences ($p < 0.05$) in scores between the baseline (i.e., ``pure'') and the impure representations. Furthermore, we use $\uparrow$/$\downarrow$ to indicate that a metric is better if its score is higher/lower and compute all metrics over 5 folds. For statistical significance validation, we include $p$ values (two-sided T-test).}
    \label{table:metric_eval}
\end{table*}

\paragraph{Baselines and Setup} We compare the purity of concept representations in various methods using our metrics.
We select representative methods from (i) \textit{supervised CL} (i.e., jointly-trained CBMs~\citep{Koh2020} with sigmoidal and logits bottlenecks, and CW~\citep{Chen2020} both when its representations are reduced through a MaxPool-Mean reduction and when no feature map reduction is applied), (ii) \textit{unsupervised CL} (i.e., CCD~\citep{Been2020} and SENN~\citep{alvarez2018towards}), (iii) \textit{unsupervised DGL} (vanilla VAE \citep{Kingma2014} and $\beta$-VAE~\citep{Higgins2017}), and (iv) \textit{weakly supervised DGL} (Ada-GVAE and Ada-MLVAE~\citep{Locatello2020}). For each method and metric, we report the average metric values and $95\%$ confidence intervals obtained from 5 different random seeds. We include details on training and architecture hyperparameters in Appendix~\ref{appendix:model_architectures}.

\subsection{Results and Discussion}
\paragraph{In contrast to DGL metrics, our metrics can meaningfully capture impurities concealed in representations.} We begin by empirically showing that our metrics indeed capture impurities encoded within a concept representation. For this, we prepare a simple synthetic dataset of ground-truth concept vectors $\mathcal{D} := \{\mathbf{c}^{(i)} \in \{0, 1\}^5\}_{i=1}^{3,000}$ where, for each sample $\mathbf{c}^{(i)}$, its $j$-th concept is a binary indicator $\mathds{1}_{\tilde{c}^{(i)}_j \ge 0}$ of the sign taken by a latent variable $\tilde{c}^{(i)}_j$ sampled from a joint normal distribution $\tilde{\mathbf{c}}^{(i)} \sim \mathcal{N}(\mathbf{0}, \Sigma)$ (with $\tilde{\mathbf{c}}^{(i)} \in \mathbb{R}^5$). During construction, we simulate real-world co-dependencies between different concepts by setting $\Sigma$'s non-diagonal entries to $0.25$. To evaluate whether our metrics can discriminate between different levels of impurities encoded in different concept representations, we construct two sets of soft concept representations. First, we construct a baseline ``\textit{pure}'' fuzzy representation $\mathbf{\hat{c}}^{(\text{pure})}$ of vector $\mathbf{c} \in \mathcal{D}$ by sampling $\hat{c}^{(\text{pure})}_j$ from $\text{Unif}(0.95, 1)$ if $c_j = 1$ or from $\text{Unif}(0, 0.05)$ if $c_j = 0$. Notice that each dimension of this representation preserves enough information to \textit{perfectly predict} each concept's activation state without encoding any extra information. In contrast, we construct a perfectly ``\textit{impure}'' soft concept representation $\mathbf{\hat{c}}^{(\text{impure})}$ by encoding, as part of each concept's fuzzy representation, the state of all other concepts. For this we partition and tile the sets $[0.0, 0.05]$ and $[0.95, 1.0]$ into $2^{5 - 1} = 16$ equally sized and disjoint subsets $\{[\text{off}_{i}, \text{off}_{i+1})\}_{i = 0}^{15}$ and $\{[\text{on}_{i}, \text{on}_{i+1})\}_{i = 0}^{15}$, respectively. From here, we generate an impure representation of ground truth concept vector $\mathbf{c} \in \mathcal{D}$ by sampling $\hat{c}_j^{(\text{impure})}$ from $\text{Unif}(\text{on}_{\text{bin}(\mathbf{c}_{-j})}, \text{on}_{\text{bin}(\mathbf{c}_{-j}) + 1})$ if $c_j = 1$ or from $\text{Unif}(\text{off}_{\text{bin}(\mathbf{c}_{-j})}, \text{off}_{\text{bin}(\mathbf{c}_{-j}) + 1})$ otherwise, where we use $\text{bin}(\mathbf{c}_{-j})$ to represent the decimal representation of the vector resulting from removing the $j$-th dimension of $\mathbf{c}$. Intuitively, each concept in this soft representation encodes the activation state of every other concept using different confidence ranges. Therefore, one can perfectly predict all concepts from a single concept's representation, an impossibility from ground truth concepts alone.

We hypothesize that if a metric is capable of accurately capturing undesired impurities within concept representations, then it should generate vastly different scores for the two representation sets constructed above. To verify this hypothesis, we evaluate our metrics, together with a selection of DGL disentanglement metrics, and show our results in Table~\ref{table:metric_eval}. We include \textit{SAP}~\cite{kumar2017variational}, $R^4$~\cite{ross2021benchmarks}, \textit{mutual information gap} (MIG)~\cite{chen2018isolating}, and \textit{DCI Disentanglement} (DCI Dis)~\cite{Eastwood2018} as representative DGL metrics given their wide use in the DGL literature~\cite{ross2021benchmarks, zaidi2020measuring}. Our results show that our metrics correctly capture the difference in impurity between the two representation sets in a statistically significant manner. In contrast, existing DGL metrics are incapable of clearly discriminating between these two impurity extremes, with DCI being the only metric that generates some statistically significant differences albeit the scores' differences are minimal (less than 0.5\%). Surprisingly, although MIG is inspired on a similar mutual information (MI) argument as our OIS metric, it was unable to capture any meaningful differences between our two representation types. We believe that this is due to the fact that in order to compute the MIG one requires an estimation of the MI which, being sensitive to hyperparameters, may fail to capture important differences. These results, therefore, support using a non-linear model's test AUC as a proxy of the MI. Further details can be found in Appendix~\ref{appendix:metric_eval}.

\begin{figure*}[h!]
    \centering
    \begin{subfigure}[b]{0.49\textwidth}
        \includegraphics[width=0.9\textwidth]{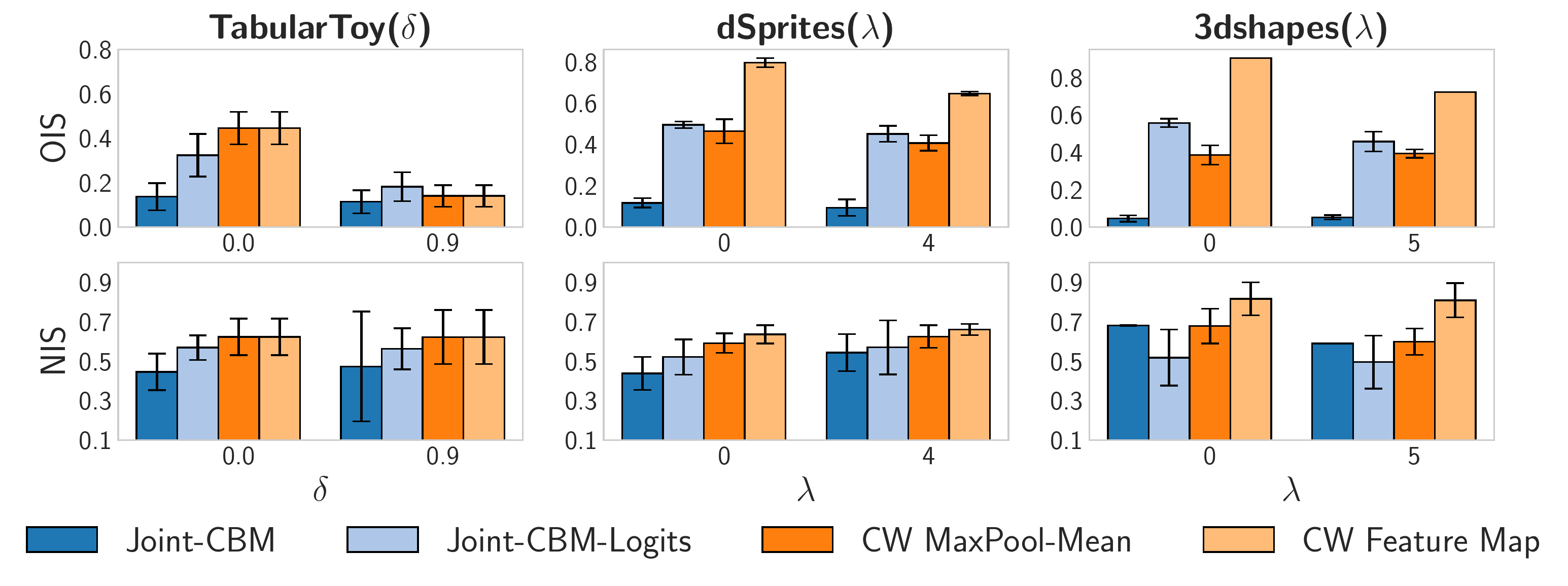}
        \caption{}
        \label{fig:representation_reduction_impurities}
    \end{subfigure}
    \begin{subfigure}[b]{0.37\textwidth}
        \centering
        \includegraphics[width=0.65\textwidth]{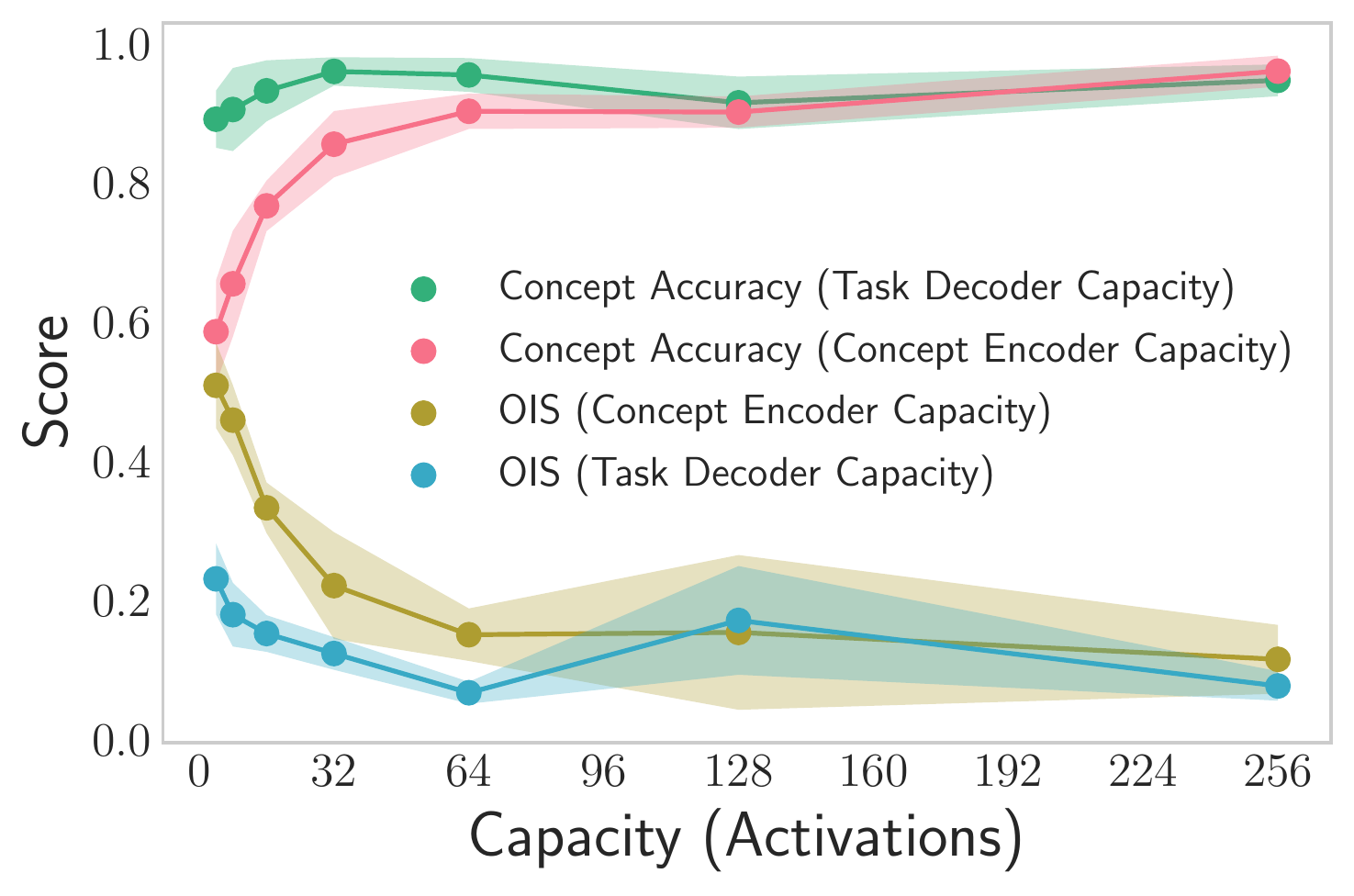}
        \caption{}
        \label{fig:toy_network_capacity}
    \end{subfigure}
    \caption{(a) Impurity scores, and their 95\% confidence intervals, for concept representations in CBM and CW (in low and high concept inter-dependence) and a corresponding transformation which leads to higher impurities. (b) Effect of network capacity (i.e., number of hidden activations used in the concept encoder and label predictor) on a CBM's concept accuracy and OIS. }
\end{figure*}

\paragraph{Our metrics can capture impurities caused by differences in concept representations and model capacities, as well as by accidental spurious correlations.}
Impurities in a CL model can come from different sources, such as differences in concept representations, as previously shown in Figure~\ref{fig:SameAccuracy}, or architectural constraints (e.g., a CBM trained with a partial/incomplete set of concepts). Here, we show that impurities caused by differences in the nature of concept representations, as well as by inadequate model capacities and spurious correlations, can be successfully captured by our metrics and thus avoided.

\textit{Differences in concept representations}: in Figure~\ref{fig:representation_reduction_impurities}, we show the impurities in (1) a CBM with a sigmoidal bottleneck (\textit{CBM}) vs a CBM with logits in its bottleneck (\textit{CBM-Logits}) and (2) a \textit{CW} module with and without feature map reduction (\textit{CW Feature Map} vs \textit{CW Max-Pool-Mean}). Our metrics show that CBM-Logits and CW Feature Map are prone to encoding more impurities than their counterparts. This is because their representations are less constrained, as logit activations can be within any range (as opposed to $[0, 1]$ in \textit{Joint-CBM}) and \textit{CW Feature Map} preserves all information from its concept feature map by not reducing it to a scalar. The exception to this is the failure of NIS to detect impurities in CBM-Logits for 3dshapes$(\lambda)$. We hypothesise that this is due to this task's higher complexity, forcing both CBM and CBM-Logits to distribute inter-concept information across all representations more than in other datasets.

\textit{Differences in model capacity}: low-capacity models may be forced to use their concept representations to encode information outside their corresponding ground truth concept. To verify this, we train a CBM in TabularToy($\delta = 0$) whose concept encoder and label predictor are three-layered ReLU MLPs. We vary the capacities of the concept encoder or label predictor by setting their hidden layers' activations to $\{\text{capacity}, \text{capacity}/2 \}$, while fixing the number of hidden units in their corresponding counterpart to $\{128, 64 \}$. We then monitor the accuracy of concept representations w.r.t.\ their aligned ground truth concepts as well as their OIS. We observe (Figure~\ref{fig:toy_network_capacity}) that as the concept encoder and label predictor capacities decrease, the CBM exhibits significantly higher impurity and lower concept accuracy. Note that the concept encoder capacity has a significantly greater effect on the purity of the representations compared to the label predictor capacity. Measuring impurities in a systematic way using our metrics can therefore guide the design of CL architectures that are less prone to  impurities.

\begin{figure}[h!]
    \centering
    \begin{subfigure}[b]{0.8\columnwidth}
        \centering
        \includegraphics[width=\textwidth]{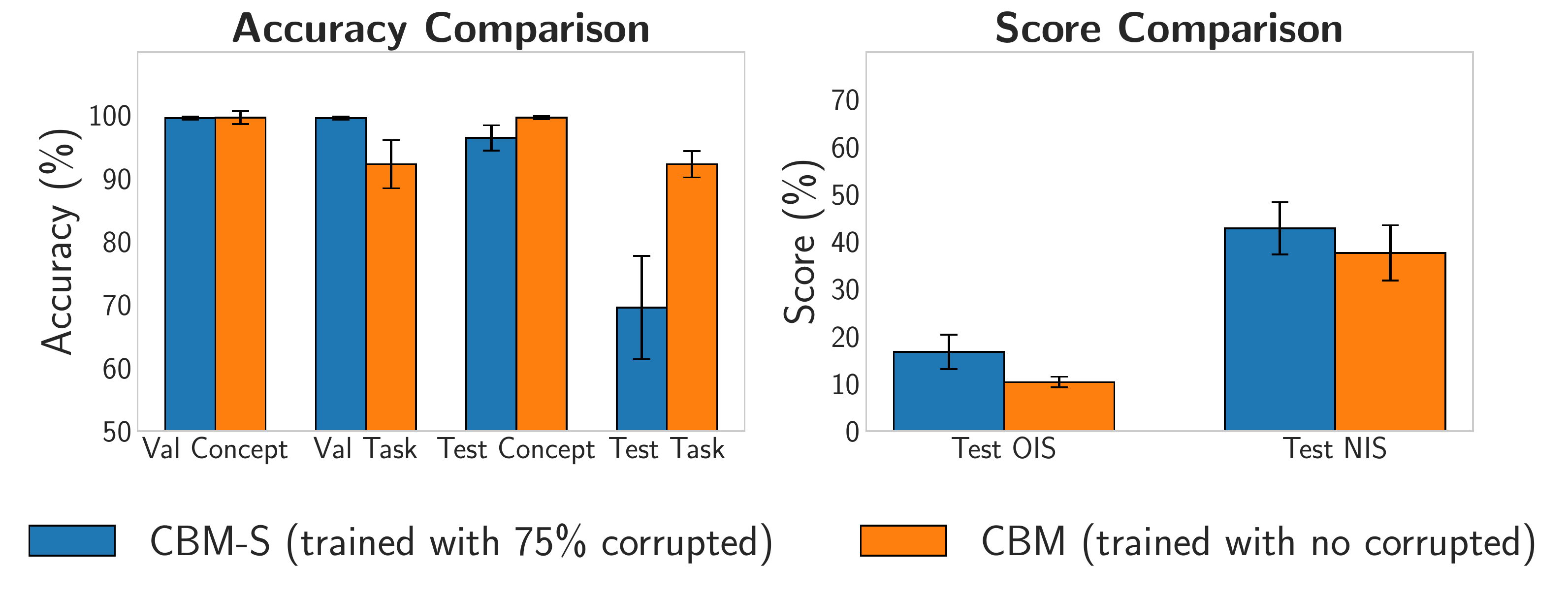}
    \end{subfigure}
    \caption{Impurity scores and validation/testing accuracies for CBMs trained on the original and spuriously corrupted dSprites$(\lambda =0)$.}
    \label{fig:SpuriousDspritesResults}
\end{figure}

\textit{Spurious correlations}: we create a variation of dSprites$(\lambda = 0)$, where we randomly introduce spurious correlations by assigning each sample a class-specific background colour with 75\% probability (see Appendix~\ref{appendix:SpuriousDsprites} for details). We train two identical CBMs on dSprites$(\lambda = 0)$ and its corrupted counterpart. During training, we note that CBM-S (the CBM trained on the corrupted data) has a higher task validation accuracy than the other CBM (Figure~\ref{fig:SpuriousDspritesResults}), while having similar concept validation accuracies. Nevertheless, when we evaluate both models using a test set sampled from the original dSprites$(\lambda = 0)$ dataset, we see an interesting result: both models can predict ground truth concept labels with similarly high accuracy. However, unlike CBM, CBM-S struggles to predict the task labels. Failure of CBM-S to accurately predict task labels is remarkable as labels in this dataset are uniquely a function of their corresponding concept annotations and CBM-S is able to accurately predict concepts in the original dSprites$(\lambda = 0)$ dataset. We conjecture that this is due to the fact that concepts in CBM-S encode significantly more information than needed, essentially encoding the background colour in addition to the original concepts as part of their representations. To verify this, we evaluate the OIS and NIS of the concept representations learnt by both models and observe that, in line with our intuition, CBM-S indeed encodes significantly more impurity. Our metrics can therefore expose spurious correlations captured by CL methods which appear to be highly predictive of concept labels while underperforming in their downstream task.

\begin{figure*}[t!]
    \centering
    \begin{subfigure}[b]{0.49\textwidth}
        \centering
        \includegraphics[width=0.98\textwidth]{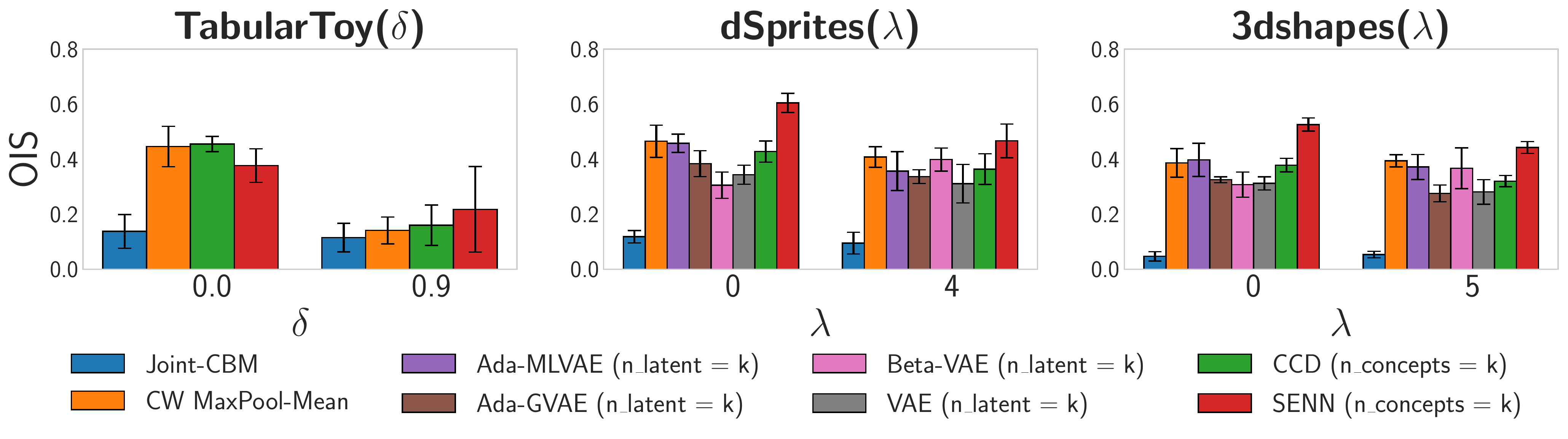}
        \subcaption{}
        \label{fig:ois_metric_results}
    \end{subfigure}
    \begin{subfigure}[b]{0.49\textwidth}
        \centering
        \includegraphics[width=0.98\textwidth]{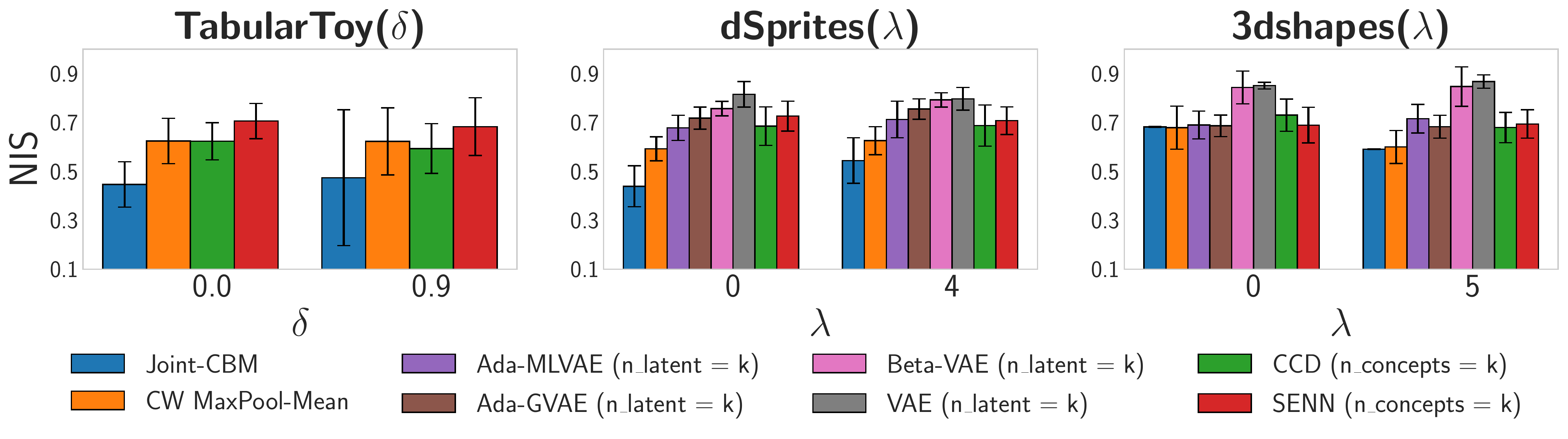}
        \subcaption{}
        \label{fig:nis_metric_results}
    \end{subfigure}
    \caption{Evaluation of different models, deployed across various tasks, using (a) our Oracle Impurity Score (OIS) metric and (b) Niche Impurity Score (NIS) metric in two extreme cases of no correlation and high concept correlation for each dataset.}
\end{figure*}

\begin{figure}[!h]
    \centering
    \includegraphics[width=0.9\columnwidth]{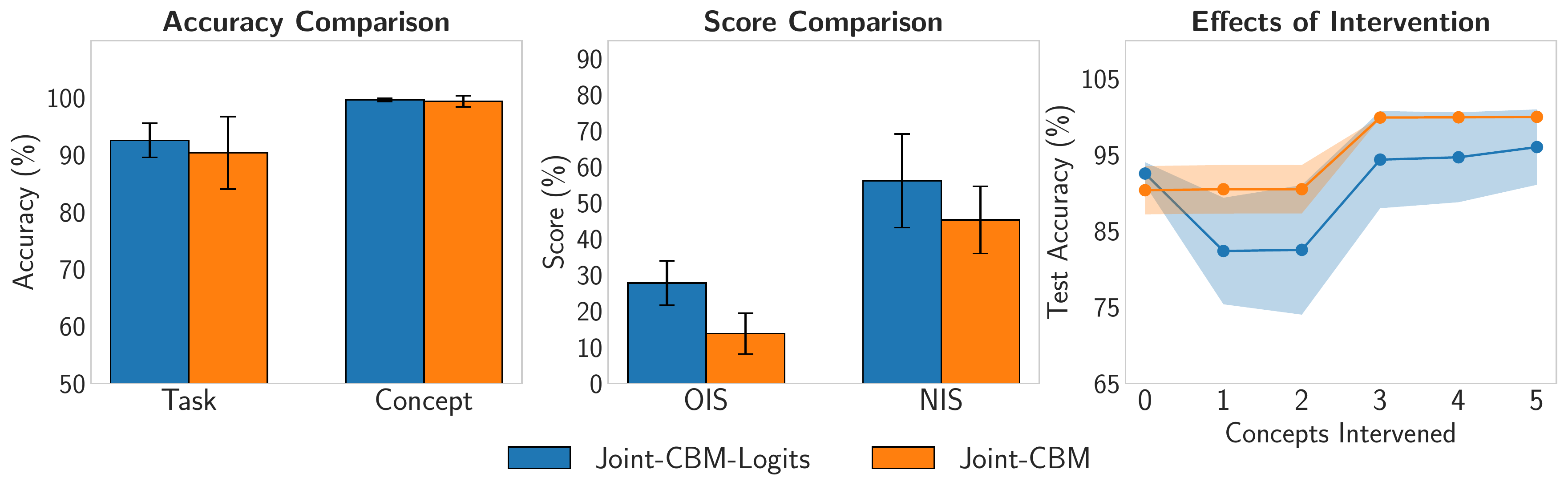}
    \caption{Intervention performance in two CBMs trained on dSprites$(\lambda = 0)$ when one uses sigmoidal concept representations and the other does not.}
    \label{fig:cbm_intervention_mixed}
    \vspace{-0.07cm}
\end{figure}

\paragraph{Our metrics can indicate when it is safe to perform interventions on a CBM by giving a realistic picture of impurities.} A major potential consequence of not being able to measure the impurities faithfully is that \textit{concept interventions}~\citep{Koh2020}, which allow domain experts to adjust the model by intervening and fixing predictions at the concept level, may fail: adjusting a concept $\mathbf{\hat{c}}_{(:, i)}$ may unintentionally impact the label predictor's  understanding of another concept $\mathbf{\hat{c}}_{(: j)}$ if representation $\mathbf{\hat{c}}_{(:, i)}$ encodes unnecessary information about concept $c_j$.
To see this in practice, consider a CBM model and a CBM-Logits model both trained to convergence on dSprites$(\lambda = 0)$, and both achieving fairly similar task and concept accuracies (Figure~\ref{fig:cbm_intervention_mixed}). We then perform interventions at random on their concept representations as follows: in CBM, where concept activations represent probabilities, we intervene on the $i$-th concept representation by setting $\hat{c}_i$ to the value of its corresponding ground truth concept $c_i$. In CBM-Logits, as in~\cite{Koh2020}, we intervene on the $i$-th concept by setting it to the 5\%-percentile of the empirical distribution of $\hat{c}_i$ if $c_i = 0$, and we set it to the 95\%-percentile of that concept's distribution if $c_i = 1$. Interestingly, our results  (Figure~\ref{fig:cbm_intervention_mixed}) show that random interventions cause a significant drop in task accuracy of CBM-Logits while leading to an increase in accuracy in CBM. Looking at the impurities of these two models, we observe that although the CBM-Logits model has a better accuracy, both of its OIS and NIS scores are considerably higher than those for the CBM model, explaining why interventions had such undesired consequences.

To rule out the difference in intervention mechanism as the cause of these results, we train two CBM-Logits with the same concept encoder capacities but with different capacities in their label predictors and observe the same phenomena as above: performance degradation upon intervention, which occurs in the case of the lower capacity model, is evidence for higher OIS and NIS scores compared to that of the higher capacity model. Further details about this experiment are documented in Appendix~\ref{appendix:intervention}.

\paragraph{Our metrics can provide insights on the impact of different degrees of supervision on concept purity.}
As models of various families benefit from varying degrees of supervision ranging from explicit supervision (supervised CL) to implicit (unsupervised CL), weak (weakly-supervised DGL) and no supervision at all (unsupervised DGL), different models are expected to learn concept representations of varying purity. The intuitive assumption is that more supervision leads to better and purer concepts. Here, we compare models from all families using our metrics and show that, contrary to our intuition, this is not necessarily the case. Within variants of CBM and CW, we choose CBM without logits and CW MaxPool-Mean, as they tend to encode fewer impurities (see Figure~\ref{fig:representation_reduction_impurities}). Furthermore, given the tabular nature of TabularToy$(\delta)$, we do not compare DGL methods in this task. Finally, for details on computing our metrics when an alignment between ground truth concepts and learnt representations is missing, see Appendix~\ref{appendix:max_alignment}.

In terms of task accuracy, the overall set of learnt concept representations is equally predictive of the downstream task across all surveyed methods (see Appendix~\ref{appendix:downstream_task_AUC} for details). However, as discussed previously, models with the same task accuracy can encode highly varied levels of impurities in their individual concept representations. Figure~\ref{fig:ois_metric_results} shows a comparison of impurities observed across methods using OIS. CBM's individual concepts consistently experience the least amount of impurity due to receiving explicit supervision, which is to be expected. Unexpectedly though, we observe that the same explicit supervision can lead to highly impure representations in CW. Indeed, CW impurities are on par or more than those of unsupervised approaches. Looking into implicit supervision, we observe that individual concepts in CCD and SENN do not correspond well to the ground truth ones. This indicates that the information about each ground truth concept is distributed across the overall representation rather than localised to individual concepts, leading to relatively high OIS. We attribute CCD's lower impurity, compared with SENN, to the use of a regularisation term that encourages coherence between concept representations in similar samples and misalignment between concept representations in dissimilar samples. More interestingly, however, both CCD and SENN encode higher impurities than some DGL approaches, despite benefiting from task supervision. Within DGL approaches, astonishingly no supervision in unsupervised DGL can result in purer individual concept representations than those of weakly-supervised DGL methods. This suggests that concept information may be heavily distributed in weakly-supervised DGL methods.

Moving from individual concepts, Figure~\ref{fig:nis_metric_results} shows a comparison of impurities observed across subsets of concept representations using our NIS metric. Similar to our OIS results, the overall set of concept representations in CBM shows the least amount of impurity. Unlike individual impurities, however, the overall set of concept representations in DGL methods shows a higher impurity than that of explicitly supervised approaches. This can be explained by the fact that DGL methods seem to learn representations that are not fully aligned with our defined set of ground truth concepts, yet when taken as a whole they are still highly predictive of individual concepts. This would lead to complement niches being highly predictive of individual ground truth concepts even when individual representations in those niches are not fully predictive of that concept itself, resulting in relatively high NIS scores and lower OIS score. Furthermore, notice that weakly supervised DGL methods show a lesser niche impurity than unsupervised DGL methods, suggesting, as in~\citet{Locatello2020}, that weakly-supervised representations are indeed more disentangled. We notice, however, that this decrease in impurity for weakly-supervised methods comes at the cost of their latent codes being less effective at predicting individual concepts than unsupervised latent codes (see Appendix~\ref{appendix:avg_concept_task_AUC}). Finally, within methods benefiting from explicit supervision, the overall set of learnt concepts in CCD has fewer impurities than that of SENN, which was similarly observed with individual concepts above.

\textbf{Our metrics are robust to concept correlations.} \hspace{2pt} 
As seen in Figures~\ref{fig:ois_metric_results} and~\ref{fig:nis_metric_results}, the preserved method ranking using our metrics in settings with and without correlations confirms our metrics' robustness to concept correlations.
\section{Conclusion}
\label{sec:conclusion}
Impurities in concept representations can lead to models accidentally capturing spurious correlations and can be indicative of potentially unexpected behaviour during concept interventions, which is crucial given that performing safe interventions is one of the main motivations behind CBMs. Despite this importance, current metrics in CL literature and the related field of DGL fail to fully capture such impurities. In this paper, we address these limitations by introducing two novel robust metrics that are able to circumvent several limitations in existing metrics and correctly capture impurities in learnt concept representations. Indeed, for the first time we are able to compare the purity of concepts in CL and DGL systematically, and show that, contrary to the common assumption, more explicit supervision does not necessarily translate to better concept quality, as measured by purity. More importantly, beyond comparison, our experiments show the utility of these metrics in designing and training more reliable and robust concept learning models. Therefore, we envision them to be an integral part of future tools developed for the safe deployment of concept-based models in real-world scenarios.

\section*{Acknowledgements}
The authors would like to thank our reviewers for their insightful comments on earlier versions of this manuscript. MEZ acknowledges support from the Gates Cambridge Trust via a Gates Cambridge Scholarship. PB acknowledges support from the European Union's Horizon 2020 research and innovation programme under grant agreement No 848077. UB acknowledges support from DeepMind and the Leverhulme Trust via the Leverhulme Centre for the Future of Intelligence (CFI), and from a JP Morgan Chase AI PhD Fellowship. AW acknowledges support from a Turing AI Fellowship under grant EP/V025279/1, The Alan Turing Institute, and the Leverhulme Trust via CFI. MJ is supported by the EPSRC grant EP/T019603/1.

\bibliography{9149.EspinosaZarlengaM}

\setcounter{table}{0}
\renewcommand{\thetable}{A\arabic{table}}
\clearpage
\appendix

\counterwithin{figure}{section}
\section{Appendix}
\subsection{Metrics Related to Properties of Concept Representations w.r.t.\ Ground Truth Concepts}
\label{appendix:metricReview}

Refer to Table~\ref{tab:metrics} for a summary of some metrics used for measuring the different properties of disentangled representations which are applicable to DGL concept representations.

\begin{table*}[!hb]
    \centering
    \begin{tabular}{p{0.5\textwidth}|p{0.5\textwidth}} 
    \textbf{Individual learnt concepts w.r.t.\ ground truth concepts} &   \textbf{Overall set of learnt concepts w.r.t.\ ground truth concepts} \\ \hline
    
     Modularity~\citep{Ridgeway2018}: Whether each learnt concept corresponds to at most one ground truth one. Measured by deviation from the ideal case where each learnt concept has high mutual information with only one ground truth concept and shares no information with others. 
     
     & Explicitness~\citep{Ridgeway2018}: Whether the overall set of learnt concepts can predict each individual ground truth concept using a simple (e.g., linear) classifier. Measured by the average predictive performance of concept vector.\\ \hline
     
    Mutual information gap~\citep{chen2018isolating}: Whether learnt concepts are disentangled. Measured by averaging the gap in mutual information between the two learnt concepts that have the highest mutual information with a ground truth concept. The mutual information needs to be estimated if concept representations are continuous. This metric generalises the disentanglement scores in~\citet{Higgins2017} and \citet{Kim2018}.
    
    & DCI Informativeness~\citep{Eastwood2018}: Whether the overall concept vector can predict each ground truth concept with low prediction error. Measured by average prediction error of concept vector. \\ \hline
         
    DCI Disentanglement~\citep{Eastwood2018}: Whether each learnt concept captures at most one ground truth one. Measured by the weighted average of disentanglement degree of each learnt concept. Such degree is calculated based on entropy, where high entropy for a learnt concept shows its equal importance for all ground truth ones and therefore its low disentanglement degree. The weight is calculated based on the aggregation of relative importance of a learnt concept in predicting each ground truth concept.& \\ \hline

    Alignment~\citep{Been2020}: Whether the learnt concepts match the ground truth ones. Measured by average accuracy of predicting each ground truth concept using the learnt one that predicts it best. \\ \hline
    
   SAP~\cite{kumar2017variational}: Whether learnt concept representations are aligned to one and only one ground truth generative factor. Measured as the average spread in importance scores (computed using e.g. linear regression) between the two most predictive concept representations for each ground truth concept. This metric, however, does not naturally capture whether a concept representation is highly predictive of two or more ground truth concepts. & \\ \hline
    
    $R^4$~\cite{ross2021benchmarks}: A generalisation of the SAP score to measure whether the learnt concept representations are aligned to one and only one ground truth generative factor. It operates similarly to SAP but it (1) uses a non-linear method for computing importance scores, (2) computes the importance spread across all concept representations rather than between the top two concept representations, and (3) computes concept representation importance scores using a \textit{bijective correspondence} (where a concept representation's importance score is a function of how predictive its representation is for a ground truth concept and how predictive that same ground truth concept is for that concept's latent representation). & \\ \hline
    
    \end{tabular}
    \caption{Metrics related to properties of concepts w.r.t.\ ground truth concepts  divided to two categories: those that capture properties of individual concepts and those that capture properties of the overall set of learnt concepts w.r.t.\ ground truth concepts. Although metrics in each column serve the same purpose, they are mathematically distinct.}
    \label{tab:metrics}
\end{table*}

\subsection{Toy CBM Correlation Experiment Details}
\label{appendix:correlation_toy_experiment}

For our toy CBM experiment described in Figure~\ref{fig:SameAccuracy} in Section~\ref{sec:background}, we consider the toy dataset proposed in~\citet{Mahin2021}, where samples have three concept annotations with no correlation and a label which is a function of the concepts (see Appendix~\ref{appendix:dataset_details} for details). Using the task and concept labels in this dataset, we train two CBM with the exact same architecture (the same architecture used for our ToyTabular dataset as described in Appendix~\ref{appendix:model_architectures}) with the exception that one CBM uses the raw logits outputted by the concept encoder in its bottleneck and the other CBM applies a sigmoidal activation to the logits outputted by the concept encoder. After training to convergence, we observe that both of these models are able to achieve an almost identical average concept accuracy and an almost perfect task accuracy. Nevertheless, as seen in Figure~\ref{fig:SameAccuracy}, when looking at the absolute value of the Pearson correlation coefficients of inter-concept representations for both models, the CBM logits model learns concept representations which are not fully independent of each other.

\subsection{Purity Matrix Implementation Details}
\label{appendix:oracle_impurity_details}
We compute the (i, j)-th entry of the purity matrix as follows: we split the original testing data $(X_\text{test}, Y_\text{test}, C_\text{test})$ into two disjoint sets, a new training set $(X_\text{train}^\prime, Y_\text{train}^\prime, C_\text{train}^\prime)$ and a new testing set $(X_\text{test}^\prime, Y_\text{test}^\prime, C_\text{test}^\prime)$, using a traditional 80\%-20\% split. We then use the concept representations learnt for the $i$-th concept for samples in $X_\text{train}^\prime$ to train a two-layered ReLU MLP $\psi_{i, j}(\cdot)$ with a single hidden layer with 32 activations to predict the truth value of the $j$-th ground-truth concept from the $i$-th learnt concept representation. In other words, we train $\psi_{i, j}(\cdot)$ using labelled samples $\Big\{ \Big( g \big( \phi(\mathbf{x}^{(l)}) \big)_{(:, i)}, \mathbf{c}_j^{(l)} \Big) \; | \; \mathbf{x}^{(l)} \in X_\text{train}^\prime \wedge \mathbf{c}^{(l)} \in C_\text{train}^\prime \big\}$. For simplicity, we train each helper model $\psi_{i, j}(\cdot)$ for $25$ epochs using batches of $128$ training samples from $X_\text{train}^\prime$.  Finally, we set the ($i$, $j$)-th entry of the purity matrix as the AUC achieved when evaluating $\psi_{i, j}(\cdot)$ on the new testing set $\Big( g \big(\phi(X_\text{test}^\prime) \big), C_\text{test}^\prime\Big)$.

\subsection{Architecture Choice for Helper Model in OIS and NIS}
\label{appendix:metric_model_ablation}

Both the OIS and NIS depend on using a separate helper parametric model in order to compute their values. In the case of the OIS, its computation requires the training of $\mathcal{O}(k^2)$ models $\psi_{i, j}(\cdot)$ which, as discussed in Appendix~\ref{appendix:oracle_impurity_details}, we implement as a simple two-layer ReLU MLP with 32 activations in its hidden layer. Similarly, to compute the NIS we need to train a helper classifier $f(\cdot)$ which, for simplicity, we implement as a simple ReLU MLP with hidden layers sizes $\{20, 20\}$. This strategy, i.e., using a helper classifier to compute a metric, is alike that seen in other metrics in the DGL space such as SAP scores~\cite{kumar2017variational} and $R^4$ scores~\cite{ross2021benchmarks}. Nevertheless, the requirement of including a parametric model as part of the metric's computation begs the question of how sensitive our metrics are to the choice of model for these helper architectures. Hence, in this section we address this question and empirically show that our metric's results are consistent once the capacity of the helper models is not constrained (i.e., due to extremely small hidden layers).

\begin{figure}[h!]
    \centering
    \includegraphics[width=\columnwidth]{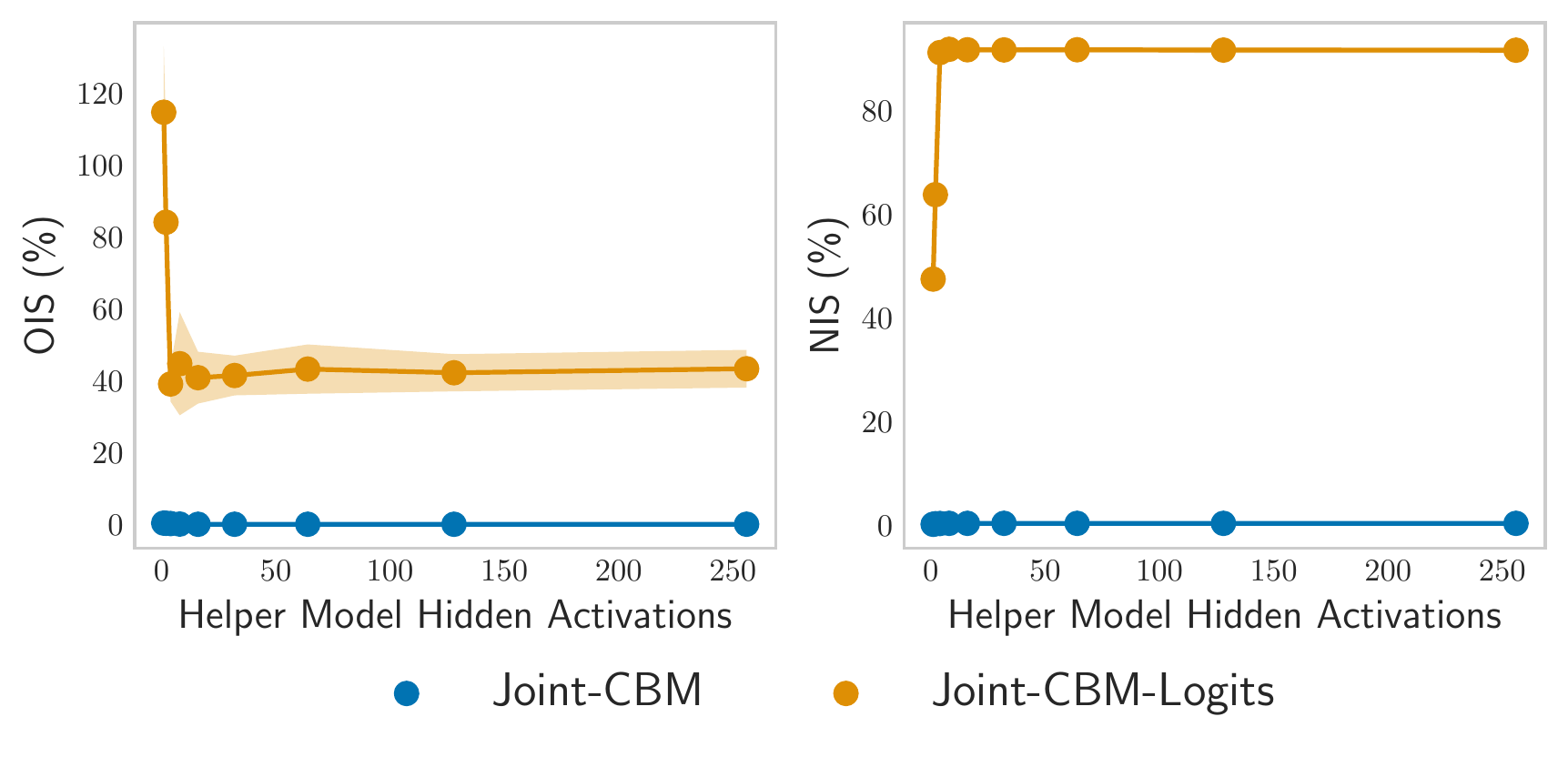}
    \caption{OIS and NIS for $\emph{dSprites}(0)$ as a function of the number of activations in the hidden layers of $\psi_{i, j}(\cdot)$ and $f(\cdot)$.}
    \label{fig:metric_architecture_ablation}
\end{figure}

In Figure~\ref{fig:metric_architecture_ablation}, we show the result of computing both the OIS and NIS for $\emph{dSprites}(0)$ as we vary the number of activations used in the hidden layers of $\psi_{i, j}(\cdot)$ and $f(\cdot)$. Specifically, we compute both metrics for a CBM with logit bottleneck activations (\textit{Joint-CBM-Logits}) and one whose activations are sigmoidal (\textit{Joint-CBM}). Both models are using the architecture and training procedure described for our other dSprites experiments in Appendix~\ref{appendix:experiment_details}. Our ablation on the capacity of the helper models shows that, for both the OIS and NIS, our metrics are stable and consistent as long as the helper models are provided with a non-trivial amount of capacity (e.g., we observe that a hidden size of at least $32$ for OIS and $16$ for NIS leads to close to constant results in dSprites). These results motivate the architectures we selected for our helper models for all the results presented in this paper and suggest that our metrics can be used in practice without the need to fine-tune their helper models.

\subsection{Concept Alignment in Unsupervised Concept Representations}
\label{appendix:max_alignment}

In the presence of ground truth concept annotations, one can still compute a purity matrix, and therefore the OIS, even when the concept representations being evaluated were learnt without direct concept supervision. We achieve this by finding an injective alignment $\mathcal{A}: \{1, 2, \cdots, k\} \mapsto \{1, 2, \cdots, k^\prime \}$ between ground truth concepts $\mathbf{c} \in \mathbb{R}^{k}$ and learnt concept representations $\mathbf{\hat{c}} \in \mathbb{R}^{d \times k^\prime}$ (we assume $k^\prime \ge k$). In this setting, we let $\mathcal{A}(i) = j$ represent the fact that the $i$-th ground truth concept $c_i$ is best represented by the $j$-th learnt concept representation $\mathbf{c}_{(:, j)}$. In our work, we greedily compute this alignment starting from a set of unmatched ground truth concepts $\mathcal{I}_\text{ground}^{(0)} = \{1, \cdots, k\}$ and a set of unmatched learnt concept representations $\mathcal{I}_\text{learnt}^{(0)} = \{1, \cdots, k^\prime\}$ and iteratively constructing $\mathcal{A}$ by adding one match $(i, j)$ at a time. Specifically, at time-step $t + 1$ we match ground truth concept $i$ with learnt concept $j$ if one can predict concept $i$ from $\mathbf{c}_{(:, j)}$ better than every other concept representation $\mathbf{c}_{(:, j^\prime)}$ can predict every other ground truth concept $c_{i^\prime}$. We evaluate predictability of ground truth concept $c_i$ from learnt concept representation $\mathbf{\hat{c}}_{(:, j)}$ by training a ReLU MLP with a single hidden layer with 32 activations and evaluating its AUC on a test set. Once a match between ground truth concept $i$ and learnt concept representation $j$ has been established, we set $\mathcal{I}_\text{ground}^{(t + 1)} := \mathcal{I}_\text{ground}^{(t)} \backslash \{i\}$  and $\mathcal{I}_\text{learnt}^{(t + 1)} := \mathcal{I}_\text{learnt}^{(t)} \backslash \{j\}$. We repeat this process until we have found a match for every ground truth concept (i.e., until $\mathcal{I}_\text{ground}^{(t)}$ becomes the empty set). Notice that in practice, one needs to compute the predictability of ground truth concept $i$ from concept representation $j$ only once when building alignment $\mathcal{A}$.

\subsection{Experiment Details}
\label{appendix:experiment_details}
In this section we provide further details on the datasets used for our experiments as well as the architectures and training regimes used in our work.

\subsubsection{Datasets}
\label{appendix:dataset_details}
For benchmarking, we use three datasets: a simple synthetic toy tabular dataset extending that defined in \citep{Mahin2021}, \emph{dSprites} \citep{matthey2017dSprites}, and \emph{3dshapes} \citep{3dshapes18}. For our evaluation, we re-implement from scratch the first dataset while we use the open-sourced MIT-licensed releases of the latter two datasets. In order to investigate the impact of concept correlation on the purity of concept representations in each method, we allow a varying degree of correlations in concept annotations and propose the following parameterized tasks:
\begin{itemize}
    \item $\emph{TabularToy}(\delta)$: this dataset consists of inputs $\{ \mathbf{x}^{(i)} \in \mathbb{R}^7 \}_{i = 1}^N$ such that the $j$-th coordinate of $\mathbf{x}^{(i)}$ is generated by applying a non-invertible nonlinear function $f_j(z_1^{(i)}, z_2^{(i)}, z_3^{(i)})$ to 3 latent factors $\{z_1^{(i)}, z_2^{(i)}, z_3^{(i)}\}$. These latent factors are sampled from a multivariate normal distribution with zero mean and a covariance matrix with $\delta$ in its off-diagonal elements. The concept annotations for each sample correspond to the binary vector $\mathbf{c}^{(i)} := [\mathds{1}(z_1^{(i)} > 0), \mathds{1}(z_2^{(i)} > 0), \mathds{1}(z_3^{(i)} > 0)]$ and the task we use is to determine whether at least two of the latent variables are positive. In other words, we set $y^{(i)}$ to $\mathds{1}\big( (c^{(i)}_1 + c^{(i)}_2 + c^{(i)}_3) \ge 2 )$. The individual functions used to generate each coordinate of $\mathbf{x}^{(i)}$ are the same as those defined in \citep{Mahin2021}. As in \citep{Mahin2021}, we use a total of $2,000$ generated samples during training and a total of $1,000$ generated samples during testing.
    \item $\emph{dSprites}(\lambda)$: we define a task based on the dSprites dataset where each sample $\mathbf{x}^{(i)} \in \{0, 1\}^{64 \times 64 \times 1}$ is a grayscale image containing a square, an ellipse, or a heart with varying positions, scales, and rotations. Each sample $\mathbf{x}^{(i)}$ is procedurally generated from a vector of ground truth factors of variation $\mathbf{z}^{(i)} = [\text{shape} \in \{0, 1, 2\}, \text{scale} \in \{0, \cdots, 5\}, \theta \in \{0, \cdots, 39\}, x \in \{0, \cdots, 31\}, y \in \{0, \cdots, 31\}]$ ($\theta$ indicating an angle of rotation) and is assigned a binary concept annotation vector $\mathbf{c}^{(i)} \in \{0, 1\}^5$ with elements $\mathbf{c}^{(i)} := [\mathds{1}\big(z^{(i)}_1 < 2\big), \mathds{1}\big(z^{(i)}_2 < 3\big), \mathds{1}\big(z^{(i)}_3  < 20 \big), \mathds{1}\big(z^{(i)}_4  < 16 \big), \mathds{1}\big(z^{(i)}_5  < 16 \big) ]$. For this task, we construct a set of $8$ labels from the concept annotations by setting $y^{(i)} = \big[c^{(i)}_2 c^{(i)}_4\big]_{10}$ if $c^{(i)}_1 = 1$ (where we use $[b_1 b_2]_{10}$ to indicate the base-10 representation of a binary number with digits $b_1$ and $b_2$) and $y^{(i)} = 4 + \big[c^{(i)}_3 c^{(i)}_5\big]_{10}$ otherwise. Finally, we parameterize this dataset on the correlation number $\lambda \in \{0, 1, \cdots, 4\}$ that indicates the number of random correlations we introduce across the sample-generating factors of variation (with $\lambda = 0$ implying all factors of variation are independent). For example, if $\lambda = 1$, we introduce a conditional correlation between factor of variations $z_1$ (``shape'') and $z_2$ (``scale'') by assigning each value of $z_1$ a random subset of values that $z_2$ may take given $z_1$. This subset is sampled by selecting, at random for each possible value of $z_1$, half of all the values that $z_2$ can take. More specifically, if $z_1$ and $z_2$ can take a total of $T_1$ and $T_2$ different values, respectively, then for each $a \in \{1, 2, \cdots, T_1 \}$ we constraint $z_2$ to be able to take only $\lfloor T_2 / 2 \rfloor$ values from the set $z_2 \in \mathcal{Z}_2(a)$ defined as:
    \[
    \mathcal{Z}_2(a) = \begin{cases} \emph{SWR}(\{1, 2, \cdots, \lfloor \frac{3 T_2}{4} \rfloor\}, \lfloor T_2 / 2 \rfloor) & c_1 = 1 \\
    \emph{SWR}(\{\lfloor \frac{T_2}{4} \rfloor, \cdots, T_2\}, \lfloor T_2 / 2 \rfloor) & \text{otherwise}
    \end{cases}
    \]
    where $\emph{SWR}(A, n)$ stands for Sample Without Replacement and is a function that takes in a set $A$ and a number $n$ and returns a set of $n$ elements sampled without replacement from $A$. This process is recursively extended for higher values of $\lambda$ by letting the dataset generated with $\lambda = i$ be the same as the dataset generated by $\lambda = i - 1$ with the addition of a new conditional correlation between factor of variations $z_{i}$ and $z_{(i + 1)}$. Finally, in order to maintain a constant dataset cardinality as $\lambda$ varies, we subsample all allowed factor of variations in $\{z_{\lambda + 2}, z_{\lambda + 3}, \cdots, z_5 \}$ by selecting only every other allowed value for each of them. This guarantees that once a conditional correlation is added, the cardinality of the resulting dataset is the same as the previous one. Because of this, all parametric variants of this dataset consist of around $\sim32,000$ samples.
    
    \item $\emph{3dshapes}(\lambda)$: we define a task based on the 3dshapes dataset where each sample $\mathbf{x}^{(i)} \in \{0, 1, \cdots, 255\}^{64 \times 64 \times 3}$ is a color image containing a sphere, a cube, a capsule, or a cylinder with varying component hues, orientation, and scale. Each sample $\mathbf{x}^{(i)}$ is procedurally generated from a vector of ground truth factors of variation $\mathbf{z}^{(i)} = [\text{floor\_hue} \in \{0, 1, \cdots, 9\}, \text{wall\_hue} \in \{0, 1, \cdots, 9\}, \text{object\_hue} \in \{0, 1, \cdots, 9\}, \text{scale} \in \{0, 1, \cdots, 7\}, \text{shape} \in \{0, 1, 2, 3\}, \text{orientation} \in \{0, 1, \cdots, 14\} ]$ and is assigned a binary concept annotation vector $\mathbf{c}^{(i)} \in \{0, 1\}^6$ with elements $\mathbf{c}^{(i)} := [\mathds{1}\big(z^{(i)}_1 < 5\big), \mathds{1}\big(z^{(i)}_2 < 5\big), \mathds{1}\big(z^{(i)}_3 < 5 \big), \mathds{1}\big(z^{(i)}_4 < 4 \big), \mathds{1}\big(z^{(i)}_5 < 2 \big), \mathds{1}\big(z^{(i)}_6 < 7 \big)]$. For this task, we construct a set of $12$ labels from the concept annotations by setting $y^{(i)} = \big[c^{(i)}_1 c^{(i)}_2 c^{(i)}_3\big]_{10}$ if $c^{(i)}_5 = 1$ and $y^{(i)} = 8 + \big[c^{(i)}_4 c^{(i)}_6\big]_{10}$ otherwise. As in the dSprites task defined above, we further parameterise this dataset with parameter $\lambda \in \{0, 1, \cdots, 5 \}$ to control the number of random conditional correlations we introduce at construction time. The procedure used to introduce such correlations is the same as in $dSprites(\lambda)$ but we use $c_5$ rather than $c_1$ for determining the set of values that we sample from. Similarly, we use the same subsampling as in $dSprites(\lambda)$ to maintain a constant-sized dataset, resulting in all parametric variants of this dataset having around $\sim16,000$ samples.
\end{itemize}

\begin{figure*}[h!]
    \centering
    \includegraphics[width=0.65\textwidth]{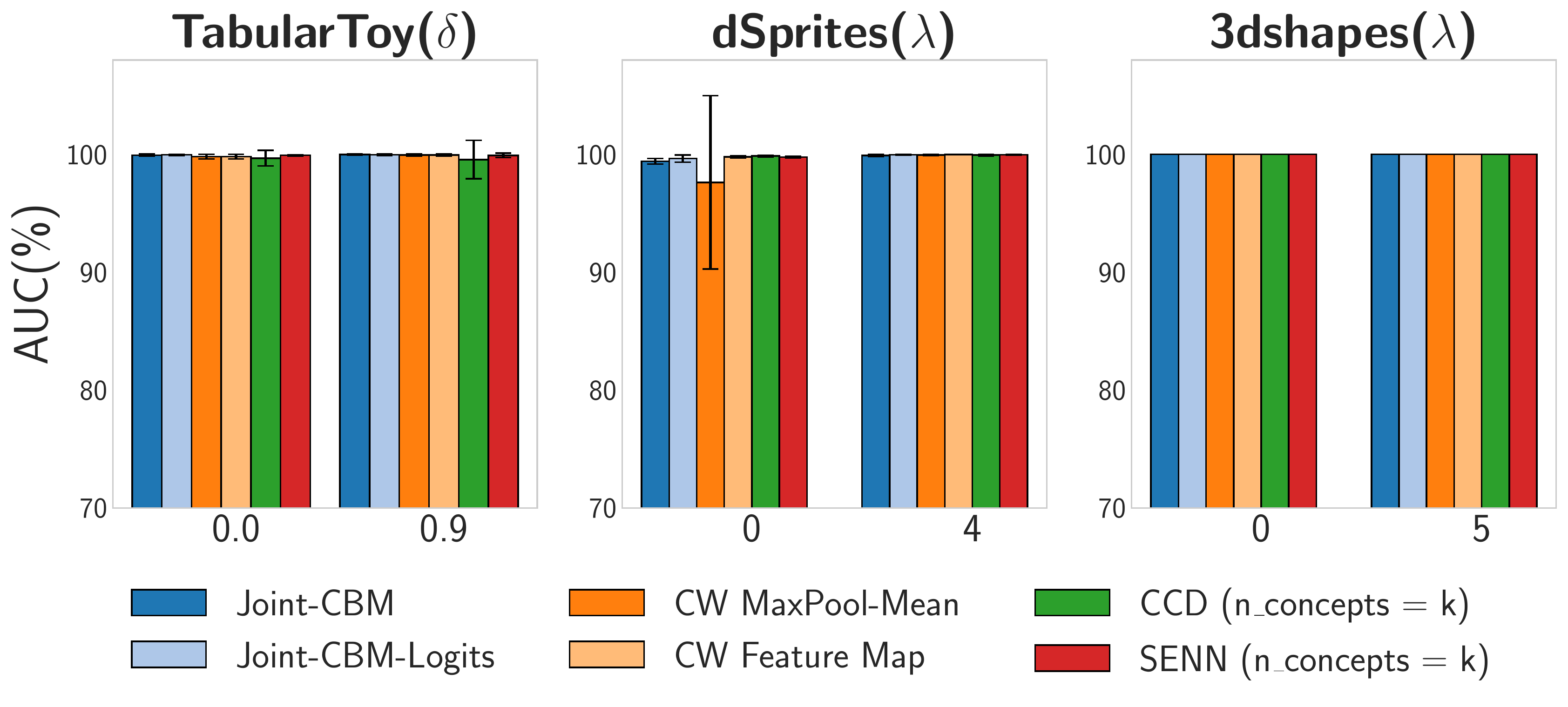}
    \caption{Downstream task predictive AUC for all datasets using original pre-trained models prior to bottleneck construction. Note that because DGL methods have no downstream task supervision in their training pipelines, we do not include those methods.}
    \label{fig:downstream_task_auc}
\end{figure*}

\subsubsection{Model Architectures and Training}
\label{appendix:model_architectures}
For our evaluation, we select CBM~\citep{Koh2020} from the supervised CL family due to its fundamental role in CL. We focus on jointly trained CBMs, where the task-specific loss and the concept prediction loss are minimised jointly.
Specifically, we evaluate CBMs that use sigmoid activations in the output of their concept encoder models (Joint-CBM) as well as CBMs with concept encoders outputting logits (Joint-CBM-Logits).
Given that CW~\citep{Chen2020} explicitly aims to decorrelate concept representations, we use it as another supervised CL baseline, both when reducing the concept representations into scalars using a MaxPool-Mean reduction as in~\citep{Chen2020} (CW-MaxPoolMean) as well as when using entire feature maps as concept representations (CW Feature Map). 
From the unsupervised CL family, CCD~\citep{Been2020} is selected due to its data agnostic nature, and SENN~\citep{alvarez2018towards} due to its particular mix of ideas from both DGL and CL literature. From the DGL family, we consider two weakly supervised methods, Adaptive Group Variational Autoencoder (Ada-GVAE) and Adaptive Multilevel Variational Autoencoder (Ada-MLVAE)~\citep{Locatello2020}, as well as two unsupervised methods, namely vanilla Variational AutoEncoders (VAE) \citep{Kingma2014} and $\beta$-VAE (with $\beta = 10$) \citep{Higgins2017}.

\paragraph{Toy Dataset Setup} With the exception of the introductory example in Section~\ref{sec:background}, in all reported  $\emph{TabularToy}(\delta)$ dataset results, for CBM, CBM-Logits, CCD, and SENN we use a 4-layer ReLU MLP with activations $[7, 128, 64, 3]$ as the concept encoder $g(\mathbf{x})$ and a 4-layer ReLU MLP with activations $[64, 128, 64, 1]$ as label predictor $f(\mathbf{\hat{c}})$ (for SENN we also include a 4-layer ReLU MLP with activations $[3, 64, 64, 7]$ as its helper decoder during training). For CW, we use the same architecture with the exception that a CW module is applied to the output of the concept encoder model. For CBM, CBM-Logits, SENN, and CCD, we train each model for $300$ epochs with a batch size of $32$. For CW, we train each model for $300$ epochs, with CW updates occurring every 20 batches, using a larger batch size of $128$ to obtain better batch statistics. Finally, for the mixture hyperparameter of the joint loss in CBM and CBM-Logits, unless specified otherwise we use a value of $\alpha = 0.1$.

For the example TabularToy experiment in Section~\ref{sec:background}, we use the same training procedure for the CBMs described above but we use a 4-layer ReLU MLP with activations $[7, 64, 64, 3]$ as the concept encoder $g(\mathbf{x})$ and a 3-layer ReLU MLP with activations $[32, 16, 1]$ as label predictor $f(\mathbf{\hat{c}})$. Everything else is set in a similar fashion as above.

\begin{figure*}[h!]
    \centering
    \includegraphics[width=0.65\textwidth]{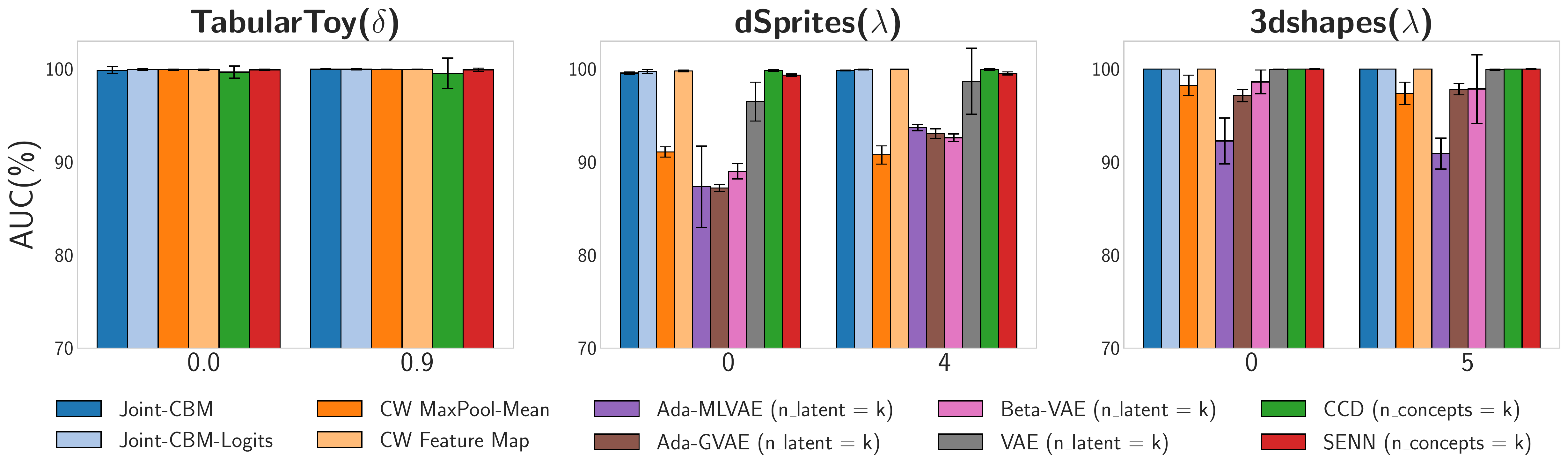}
    \caption{Downstream task predictive AUC for all datasets of a simple MLP trained to predict task labels from the overall set of learnt concepts. Note that this plot faithfully replicates the downstream task predictive AUC of methods that received direct task supervision in their training (shown in Figure~\ref{fig:downstream_task_auc}).}
    \label{fig:downstream_task_auc_from_concepts}
\end{figure*}

\subsubsection{Implementation Details}
\label{appendix:implementation_details}

\paragraph{dSprites$(\lambda)$ and 3dshapes$(\lambda)$ Setup} With the exception of our dSprites intervention experiments in Section~\ref{sec:experiments}, for our CBM, CBM-Logits, and CCD experiments in $\emph{dSprites}(\lambda)$ and $\emph{3dshapes}(\lambda)$, we use a Convolutional Neural Network (CNN) with four (3x3-Conv + BatchNorm + ReLU + MaxPool) blocks with feature maps $\{8, 16, 32, 64 \}$ followed by a three fully-connected layers with activations $\{64, 64, k\}$ for the concept encoder model $g(\mathbf{x})$ (with $k$ being the number of ground truth concepts in the dataset). Furthermore, for the label predictor model $f(\mathbf{\hat{c}})$ we use a simple 4-layer ReLU MLP with activations $\{k, 64, 64, L\}$ (with $L$ being the number of output labels in the task). For CW's concept encoder $g(\mathbf{x})$ we use a CNN with three (3x3-Conv + BatchNorm + ReLU + MaxPool) blocks with feature maps $\{8, 16, 32\}$ followed by a (3x3-Conv + CW) block with $64$ feature maps. For the label predictor, we use a model composed of a MaxPool layer followed by five ReLU fully connected layers with activations $\{64, 64, 64, 64, L\}$. All models evaluated for CBM, CBM-Logits, and CCD are trained for $100$ epochs using a batch size of $32$. In contrast, CW models are trained for $100$ epochs using a batch size of $256$ and an CW module update step every $20$ batches. Finally a value of $\alpha = 10$ is used during joint CBM and CBM-Logits training.

For the CBM and CBM-Logit models used in our intervention experiments, we use the same training setup as described above, and the same concept encoder architecture as described above, however we use different label predictor architectures. For the CBMs with the same capacity but different activation function on their concept encoders (i.e., logit vs non-logit) we use a simple 4-layer ReLU MLP with activations $\{k, 16, 8, L\}$ as their label predictor architectures. For the intervention experiment showcasing two CBMs with different capacities in Appendix~\ref{appendix:intervention}, we use the same concept encoder architecture as described for CBMs above and use a ReLU MLP with layers $\{32, 16\}$ for label predictor in the higher-capacity model while using a ReLU MLP with layers $\{8, 4\}$ for the label predictor in the lower-capacity model.

For evaluating VAE, $\beta$-VAE ($\beta = 10)$, Ada-GVAE and Ada-MLVAE, we use the same architecture as in CBM's and CCD's concept predictor for the concept encoder and the same architecture as in \citep{Locatello2020} for the decoder. The decoder consists of two ReLU fully connected layers with activations $\{ 256, 512\}$ followed by four ReLU deconvolutional layers with feature maps $\{64, 32, 32, \emph{input\_feature\_maps} \}$.  All DGL models are trained for $100$ epochs using a batch size of $32$. Weakly-supervised models are trained with a dataset consisting of $\frac{2 N}{3}$ pairs of images that share at least one factor of variation (with $N$ being the number of samples in the original dataset) while unsupervised models are trained with the same dataset used for CL methods. As in other methods, we train all DGL models using a default Adam optimizer \citep{kingma2014adam}, with learning rate $10^{-3}$.

When evaluating CCD, we use a threshold of $\beta = 0.0$ for computing concepts scores and the same regulariser parameters $\lambda_1 = 0.1$, $\lambda_2 = 0.1$, $\epsilon = 10^{-5}$ as in \citet{Been2020}'s released code for their work in \citep{Been2020}. Finally, all CCD models, across all tasks, are trained for $100$ epochs and a batch size of $32$ using a default Adam optimizer, with learning rate $10^{-3}$.

When benchmarking SENN, we use the same architecture as in DGL methods for the concept encoder $g(\mathbf{x})$ and for its corresponding decoder. Note that the decoder in this case is only used as part of the regularization term during training. For the weight model $\theta(\mathbf{\hat{c}})$ (i.e., the ``parameterizer''), we use a simple ReLU MLP with unit sizes $\{\text{input shape}, 64, 64, k\}$ ($k$ being the number of concepts SENN will learn). Finally, we use an additive aggregation function and use $\lambda = 0.1$ as a robustness regularization strength and $\zeta = 2\times10^{-5}$ as the sparsity regularization strength, as done in \cite{alvarez2018towards}. We train our SENN models for 100 epochs using a batch size of 32 and a default Adam optimizer with learning rate $10^{-3}$.

\paragraph{Libraries} We implemented all of our methods in Python 3.7 using TensorFlow~\cite{abadi2016tensorflow} as our main framework. For our implementation of CW, we adapted and modified the MIT-licensed public release of CW accompanying the original paper~\citep{Chen2020}. Similarly, for our implementation of all DGL methods, we adapted the MIT-licensed and open-sourced library accompanying the publication of~\cite{Locatello2020}. Finally, we make use of the the open-sourced MIT-licensed library released by~\citet{Kazhdan2021} for easy access to concept-based evaluation metrics and dataset wrappers for 3dshapes and dSprites. All other methods and metrics were implemented from scratch and will be made publicly available upon release of the paper.

\paragraph{Resource Utilization} We ran our experiments on a private 125Gi-RAM GPU cluster with 4 Nvidia Titan Xp GPUs and 40 Intel(R) Xeon(R) CPUs (E5-2630 v4 @ 2.20GHz). We estimate that around 180 GPU-hours were required to complete all of our experiments (this includes running all experiments across 5 random seeds).

\subsection{DGL Metric Evaluation Details}
\label{appendix:metric_eval}
In order to compare our metrics with commonly used DGL metrics in the toy dataset of  Section~\ref{sec:experiments}, we make use of the open-sourced library accompanying the publication of~\cite{Locatello2020} to compute DCI and MIG scores. Furthermore, we follow the implementation accompanying the publication of~\cite{ross2021benchmarks} for evaluating the $R^4$ metric and SAP scores. When computing the MIG, we follow the same steps as~\citet{ross2021benchmarks} and estimate the mutual information using a 2D histogram with $20$ bins. As in the rest of our experiments, we report each metric after computing it over toy datasets sampled with 5 different random seeds and include p-values computed using a two-sided Welch T-test~\cite{welch1947generalization}.

\subsection{Corrupted dSprites$(\lambda=0)$ Samples} \label{appendix:SpuriousDsprites}

\begin{figure}[h!]
    \centering
    \includegraphics[width=0.85\columnwidth]{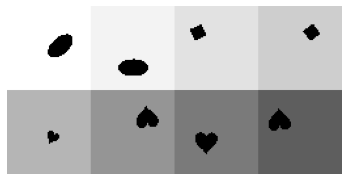}
    \caption{Different class samples 
    in dSprites$(\lambda = 0)$ with a spurious correlation introduced assigning each class a fixed background.}
    \label{fig:SpuriousDsprites}
\end{figure}

Figure~\ref{fig:SpuriousDsprites} shows 8 corrupted dSprites($\lambda = 0)$ samples, one for each class in this task, where a spurious correlation was introduced by setting the background in sample $\mathbf{x^{(i)}}$ to $0.1 \cdot y^{(i)}$, where $y^{(i)}$ is the label assigned to $\mathbf{x^{(i)}}$ in dSprites($\lambda = 0)$. In our experiments, we introduce this spurious correlation for 75\% of the training samples at random.

\subsection{Downstream Task Predictive AUC} \label{appendix:downstream_task_AUC}

Figure~\ref{fig:downstream_task_auc} shows that the downstream task predictive AUC for all datasets using raw inputs, in absence of any correlations as well as the maximum correlations between concepts, is relatively high and similar across methods. Nvertheless, if concept representations from various methods are good surrogates to the inputs, we also expect them  to be able to recover the same predictive performance. Figure~\ref{fig:downstream_task_auc_from_concepts} confirms that this holds to a good degree by looking at the task AUC of a simple ReLU MLP with hidden layers $\{ 64, 64\}$ trained to predict the corresponding task labels using the concept representations learnt by each method.

\subsection{Intervention Performance in Two CBM-Logits With Different Capacities} \label{appendix:intervention}

\begin{figure}[h!]
    \centering
    \includegraphics[width=\columnwidth]{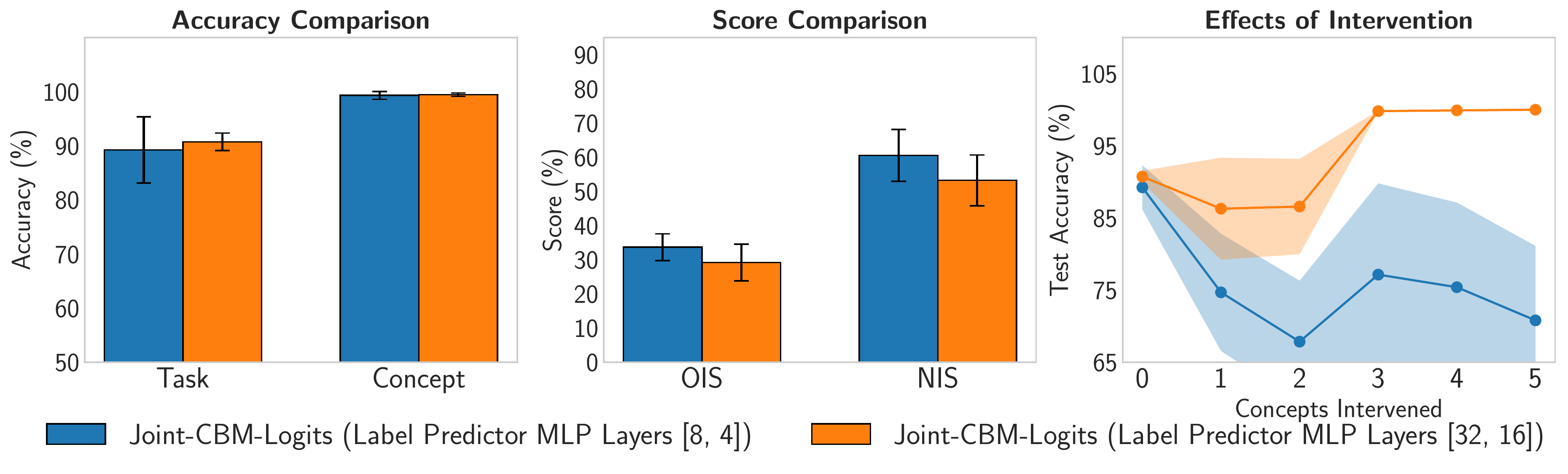}
    \caption{Intervention performance in two CBM-Logits trained on dSprites$(\lambda = 0)$ with different label predictor capacities.}
    \label{fig:cbm_intervention_logits}
\end{figure}

To discard the possibility that our intervention results presented in Section~\ref{sec:experiments} are due to the different intervention mechanisms used by Joint-CBM and Joint-CBM-Logits, in this section we show that a similar trend can be observed in CBMs using the same intervention mechanism but with different capacities. To observe this, we train two CBM-Logits with the same concept encoder capacities but with different capacities in the MLPs used as label predictors for both of these models. We see that the higher-capacity model has a higher task accuracy (by about ~4\%), but a similar concept accuracy to the lower-capacity model. Randomly intervening on their representations, however, produces very different results: interventions in the higher-capacity model lead to a boost in performance while interventions on the lower capacity model lead to performance degradation. Upon inspecting the impurities (see Figure~\ref{fig:cbm_intervention_logits}), we see that the lower-capacity model has considerably higher OIS and NIS scores than the higher-capacity model, explaining the performance degradation caused by interventions in the lower-capacity model.

\begin{figure}[h!]
    \centering
    \includegraphics[width=\columnwidth]{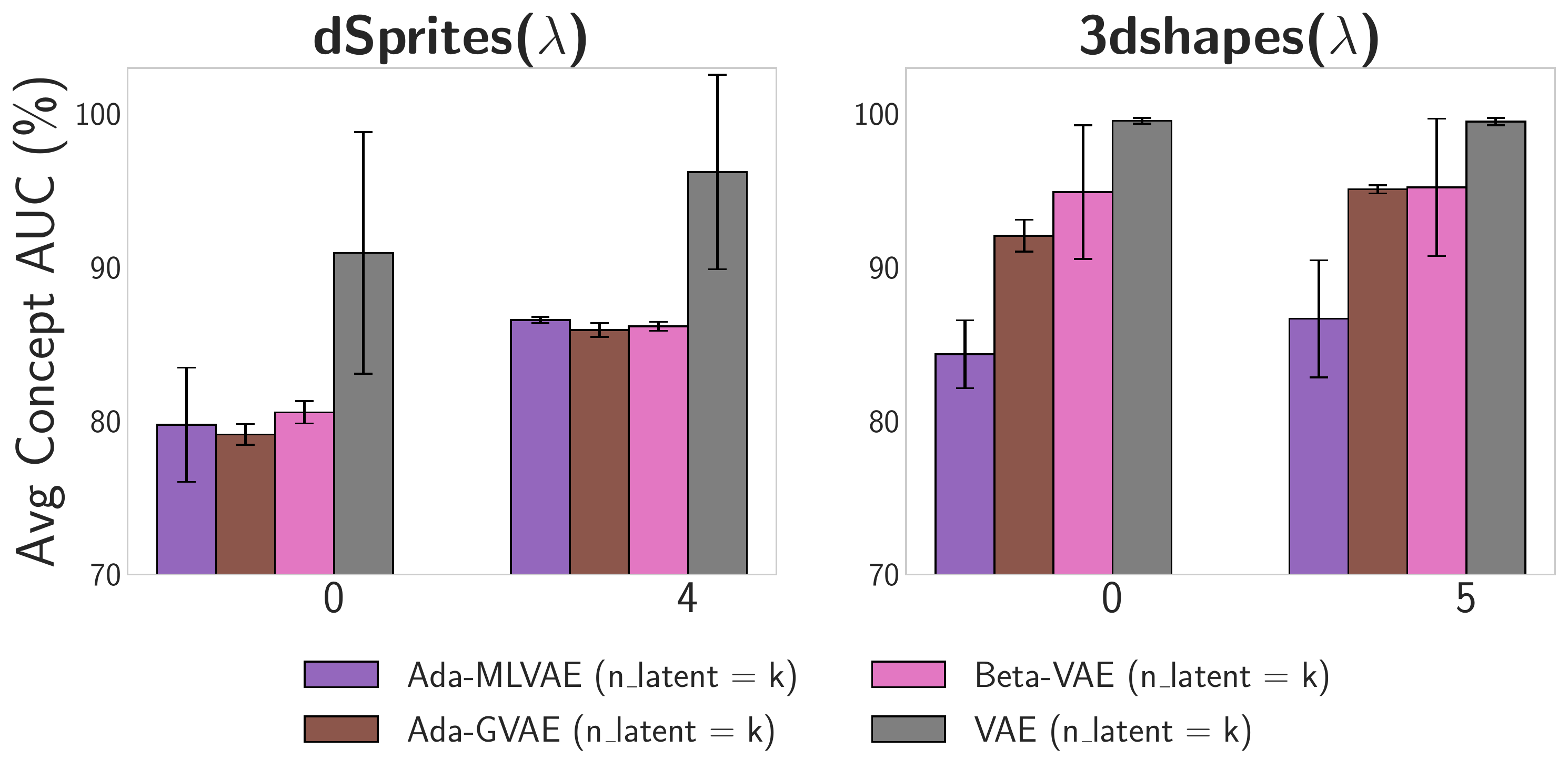}
    \caption{Mean concept AUC calculated by averaging over all the AUCs of predicting individual ground-truth concepts from the entire set of learnt concepts in DGL methods. Each AUC is calculated using a ReLU MLP with hidden layers $\{ 64, 64 \}$ that is trained to predict the target concept.
    } 
    \label{fig:avg_concept_auc_from_concepts}
\end{figure}

\subsection{Average Concept Predictive AUC from Overall Concept Representations in DGL} \label{appendix:avg_concept_task_AUC}
When benchmarking methods in Section~\ref{sec:experiments}, we observe that weakly-supervised DGL methods show a lesser niche impurity than unsupervised DGL methods. Here we hypothesise that this difference is due to their overall representations being less predictive of individual concepts than the representations learnt by unsupervised DGL methods. Figure~\ref{fig:avg_concept_auc_from_concepts} confirms this hypothesis by showing the average predictive concept AUC from a classifier trained to predict ground truth concepts from the overall concept representations of each of our DGL methods. If the overall concept representations are able to correctly capture all of the ground truth concepts, then one would expect a classifier trained on those representations to be highly predictive of all concepts. Nevertheless, notice that, as predicted, weakly-supervised DGL methods appear to have overall concept representations that are, on average, less predictive of ground truth concepts than those in unsupervised DGL methods. This result provides an explanation for why we observe a lower niche impurity score in weakly-supervised DGL methods compared to unsupervised DGL and suggests future work which could better explore the cause for this observation.

\end{document}